\documentclass[sn-mathphys]{sn-jnl}
\usepackage{diagbox}

\jyear{2023}%

\theoremstyle{thmstyleone}

\theoremstyle{thmstyletwo}%

\theoremstyle{thmstylethree}%
\newtheorem{definition}{Definition}%

\begin{document}

\title[ ]{Quantifying and Explaining Machine Learning Uncertainty in Predictive Process Monitoring: An Operations Research Perspective}

\author*[1,2]{\fnm{Nijat} \sur{Mehdiyev}}\email{nijat.mehdiyev@dfki.de}

\author[1,2]{\fnm{Maxim} \sur{Majlatow}}\email{maxim.majlatow@dfki.de}

\author[1,2]{\fnm{Peter} \sur{Fettke}}\email{peter.fettke@dfki.de}

\affil[1]{\orgname{German Research Center for Artificial Intelligence (DFKI)}, \orgaddress{\street{Campus D 3.2}, \city{Saarbrücken}, \postcode{66123}, \state{Saarland}, \country{Germany}}}

\affil[2]{\orgname{Saarland University}, \orgaddress{\street{Campus D 3.2}, \city{Saarbrücken}, \postcode{66123}, \state{Saarland}, \country{Germany}}}

\abstract{
This paper introduces a comprehensive, multi-stage machine learning methodology that effectively integrates information systems and artificial intelligence to enhance decision-making processes within the domain of operations research. The proposed framework adeptly addresses common limitations of existing solutions, such as the neglect of data-driven estimation for vital production parameters, exclusive generation of point forecasts without considering model uncertainty, and lacking explanations regarding the sources of such uncertainty. Our approach employs Quantile Regression Forests for generating interval predictions, alongside both local and global variants of SHapley Additive Explanations for the examined predictive process monitoring problem. The practical applicability of the proposed methodology is substantiated through a real-world production planning case study, emphasizing the potential of prescriptive analytics in refining decision-making procedures. This paper accentuates the imperative of addressing these challenges to fully harness the extensive and rich data resources accessible for well-informed decision-making.
}


\maketitle

\section{Introduction}
In today's highly competitive and complex business environment, organizations are under constant pressure to optimize their performance and decision-making processes. According to Herbert Simon, enhancing organizational performance relies on effectively channeling finite human attention towards critical data for decision-making, necessitating the integration of information systems (IS), artificial intelligence (AI) and operations research (OR) insights \cite{simon1997future}.  Recent OR research provides evidence in support of this proposition, as the discipline has witnessed a transformation due to the abundant availability of rich and voluminous data from various sources coupled with advances in machine learning \cite{frazzetto2019prescriptive}. As of late, heightened academic attention has been devoted to prescriptive analytics, a discipline that suggests combining the results of predictive analytics with optimization techniques in a probabilistic framework to generate responsive, automated, restricted, time-sensitive, and ideal decisions \cite{lepenioti2020prescriptive}.

The confluence of AI and OR is evident due to their interdependent and complementary nature, as both disciplines strive to augment decision-making processes through computational and mathematical methodologies \cite{boutilier2023introducing}. Various integration scenarios exist for combining these disciplines in order to address real-world problems.  A prevalent method is the "predict-then-optimize approach", where machine learning is used to predict essential parameters of an optimization model before or simultaneously as the optimization models are solved \cite{mivsic2020data}. In this context, data-driven analytics techniques have proven effective in addressing a diverse range of operations research (OR) problems, encompassing areas such as capacity planning \cite{youn2022planning}, production planning and scheduling \cite{usuga2020machine}, distribution planning \cite{kumar2020quantitative}, inventory management \cite{van2015optimization}, transportation \cite{chung2017cascading}, sales and operations planning \cite{thome2012sales}, dynamic pricing and revenue management \cite{xue2016pricing}. Employing data-driven decision-making within these business operations leads to notable benefits, including enhanced return on investment and equity, optimized asset utilization, and increased market value \cite{mehdiyev2020local}.

While the importance and adoption of data-driven solutions in operational decision-making continue to expand, it is essential to tackle certain limitations to fully realize the potential of prescriptive analytics. 
One of the significant gaps is related to the current application of supervised machine learning models, wherein they serve to estimate primarily certain non-technical parameters that are subsequently employed as either constraints or the objective function in a chosen optimization framework.
As stated by \cite{ho1989evaluating}, planning scenario uncertainties can be methodically divided into two separate categories: (i) environmental uncertainty and (ii) system uncertainty. Environmental uncertainty involves external parameters affecting the production process, such as demand and supply fluctuations, while system uncertainty includes uncertainties inherent to the production process, such as operational yield, production lead time, quality-related issues, and potential system failures, among others. An extensive review of the prescriptive analytics literature indicates that most studies focus on the first category, with technical production parameters frequently neglected \cite{chaari2014scheduling}. This oversight can be linked to the lack of relevant data from the information systems at the shop floor level \cite{mehdiyev2022deep}. Consequently, current estimates are predominantly based on intuition or assumptions and frequently yield suboptimal outcomes. 

Another considerable gap in the field pertains to the output produced during the predictive analytics stage. Upon closer examination of studies that integrate data-driven parameter estimation before optimization, it becomes apparent that they overwhelmingly produce point forecasts \cite{mitrentsis2022interpretable}. This approach, however, leads to the application of deterministic optimization methods, which may not fully capture the complexities and uncertainties of real-world scenarios. Despite the existence of numerous optimization methods that consider uncertainty, as proposed in \cite{mula2006models}, their integration with the preceding predictive analytics stage has yet to be established. Consequently, a rising demand exists for the production of machine learning outputs that can precisely and comprehensively capture and quantify the inherent uncertainty within the specific operational research context being examined.

Lastly, even if uncertainty associated with the optimization parameter of interest can be estimated, a more actionable and advantageous approach  would involve explaining its underlying source.  This can be accomplished through an Explainable Artificial Intelligence (XAI) approach that focuses on identifying input patterns that lead to uncertain predictions. By pinpointing the specific input features contributing to predictive uncertainty, practitioners can gain insights into regions where training data is sparse or where specific features exhibit anomalous behavior \cite{antoran2020getting}. These insights would alert necessary adjustments to the model's decision-making process or outcomes prior to its subsequent operationalization.

In order to address the three identified gaps, we propose a multi-stage machine learning approach that incorporates uncertainty awareness and explainability. We demonstrate the effectiveness of this approach through its application to a real-world production planning scenario. The contribution of this study is multifaceted.  To address the first gap, we utilize a supervised learning approach to probabilistically estimate a production-related parameter, specifically the processing time of production events. To achieve this objective, we employ process event data sourced from Manufacturing Execution systems (MES). These systems are process-aware information systems (PAIS) that facilitate the coordination of underlying operational processes and capture the digital footprints of process events during execution. The resulting event log consists of sequentially recorded events associated with a particular case, along with various attributes such as timestamps, resources (human or machine) responsible for process execution, and other case-specific details. To be more precise, the problem at hand is formulated as a predictive process monitoring problem. This necessitates the utilization of specific pre-processing, encoding, and feature engineering techniques to account for the inherent business and operational process data requirements.

To tackle the second gap, we utilize Quantile Regression Forests (QRF), a machine learning method developed specifically to estimate conditional quantiles for high-dimensional predictor variables \cite{Meinshausen}. QRF is an extension of the traditional random forests technique, offering a non-parametric and precise approach for estimating prediction intervals. These prediction intervals provide valuable information on the uncertainty of model outcomes, allowing for a better understanding of the predictive power of the model and its limitations.

To bridge the third gap, we offer an explanation of the main drivers of uncertainty by examining the impact of feature values on prediction intervals. To accomplish this, we utilize local and global post-hoc explanations using SHapley Additive Explanations (SHAP)  \cite{SHAP_Lundberg}. Our approach differs from the state-of-the-art use of this technique in that we use the prediction interval as the output, which provides a direct explanation of feature attributions to uncertainty. Furthermore, we refine our explanations on a granular level, such as for different uncertainty profiles or individual production activities, resulting in a more nuanced understanding of the underlying drivers of uncertainty.

The remainder of this paper is structured as follows: Section \ref{Process_Data_Use-Case_Description} presents the real-world use-case scenario under examination, which serves as the context for the demonstration of our proposed approach. Section \ref{Methodology} elucidates the essential methodology employed, while Section \ref{Experiment_Settings} introduces the experimental setup and model evaluation measures. Section \ref{Results} offers a comprehensive account of the evaluation outcomes of our proposed approach. Subsequently, Section \ref{Discussion} engages in a discussion of these findings, and Section \ref{Background_and_Related_Work} presents a synthesis of relevant literature pertaining to our proposed approach. Lastly, Section \ref{Conclusion} draws this paper to a close.

\section{Motivating Usage Scenario} \label{Process_Data_Use-Case_Description}
In this section, we delineate the pertinent production processes associated with the gathered data and explicate the practical implementation of the suggested methodology, thereby establishing a contextual foundation for orientation. It is, however, worth noting that the proposed uncertainty-aware explainable process prediction approach is amenable to transferability for diverse planning scenarios. This use case emerges from a collaborative research project involving a medium-sized German manufacturing partner specializing in custom and standardized vessel components. The multi-stage production process, employing stainless steel, aluminum, and carbon steel materials, requires specialized equipment and expertise.

At the commencement of the manufacturing process, customer orders are obtained through the partner's product catalog. Upon receiving an order, the manufacturing firm evaluates its priority and ascertains the predetermined or marginally modified sequence of production activities necessitated by the customer order specifications. These specifications encompass a variety of attributes, such as \textit{article group identifier, material group identifier, weight, bend radius, base diameter, sheet width, quantity, welding specifications}, and others that significantly influence the duration of the respective production activities. Despite possessing a sequence of activities for each customer order, process experts presently depend on intuition or experience-based estimations to ascertain their duration. This inability to quantify this vital time-specific production parameter results in suboptimal planning outcomes.

To address this issue, the partner has implemented an MES solution to capture the process execution details of production activities for each customer order. In our use case, the process data adheres to a particular structure. Each customer order is represented by a case with a unique case identifier, and a process case comprises the causal and temporal sequence of several events related to the production of the corresponding customer order. A process event encompasses the activity describing the production step executed, the start and end timestamps of execution, case attributes such as customer order specifications detailed above, and event-specific attributes like machine or human resources responsible. The examined use case involves 30 distinct activities, including \textit{forming of material on dishing presses, manual welding, plasma welding, surface grinding, manual sanding, and deburring of edges, etc.} All process execution data of historical customer orders are exported and stored in an event log. Table \ref{Table_Event_Log_Example} presents an excerpt from the event log for illustrative purposes.
\begin{table}[h]
\centering
\caption{Production Process Event Log} \label{Table_Event_Log_Example}
\begin{tabular}{|c|c|c|c|c|c|c|c|}
\hline
\textbf{Case}   & \textbf{}         & \textbf{Start}    & \textbf{End}  & \textbf{Diameter}     & \textbf{}     & \textbf{Worker}          & \textbf{} \\
\textbf{Nr}     & \textbf{Activity} & \textbf{Time}     & \textbf{Time} & \textbf{Base}         & \textbf{...}  & \textbf{ID}   & \textbf{...}\\
\hline\hline
162384          & Plasma            & 2019-04-18        & 2019-04-18    &  1800                 & ...           & 409           & \\
                & Welding           & 06:26:47          & 09:51:25      &                       &               &               & \\
\hline
162384          & Grinding          & 2019-04-18        & 2019-04-18    &  1800                 & ...           & 108           & \\
                & Weld. Seam        & 12:11:30          & 19:07:14      &                       &               &               & \\
\hline
162384          & Dishing           & 2019-04-23        & 2019-04-23     &  1800                 & ...           & 150           & \\
                & Press (2)         & 10:50:31          & 18:34:11       &                       &               &               & \\
\hline
162384          & Bead              & 2019-04-24        & 2019-04-24     &  1800                 & ...           & 726           & \\
                & Small             & 10:20:13          & 19:57:45      &                       &               &               & \\
\hline
162384          & X-Ray             & 2019-04-25        & 2019-04-25     &  1800                 & ...           & 703           & \\
                & Examination       & 10:26:23          & 10:26:32      &                       &               &               & \\
\hline
162384          & Edge             & 2019-04-26        & 2019-04-26     &  1800                 & ...           & 742           & \\
                & Deburring        & 09:08:38          & 17:50:27     &                       &               &               & \\
\hline
...             & ...               & ...               & ...           &  ...                  & ...           & ...           & ...\\
\hline
177566          & 3D Micro-         & 2021-06-21        & 2021-06-21	&  2                    & ...           & 139           & ...\\
                & step Circle        & 07:04:38          & 10:26:37      &                       & ...           &           & ...\\
\hline
...             & ...               & ...               & ...           &  ...                  & ...           & ...           & ...\\
\hline
\end{tabular}
\end{table}

\begin{figure}[h]%
\centering
\includegraphics[width = 0.9\textwidth]{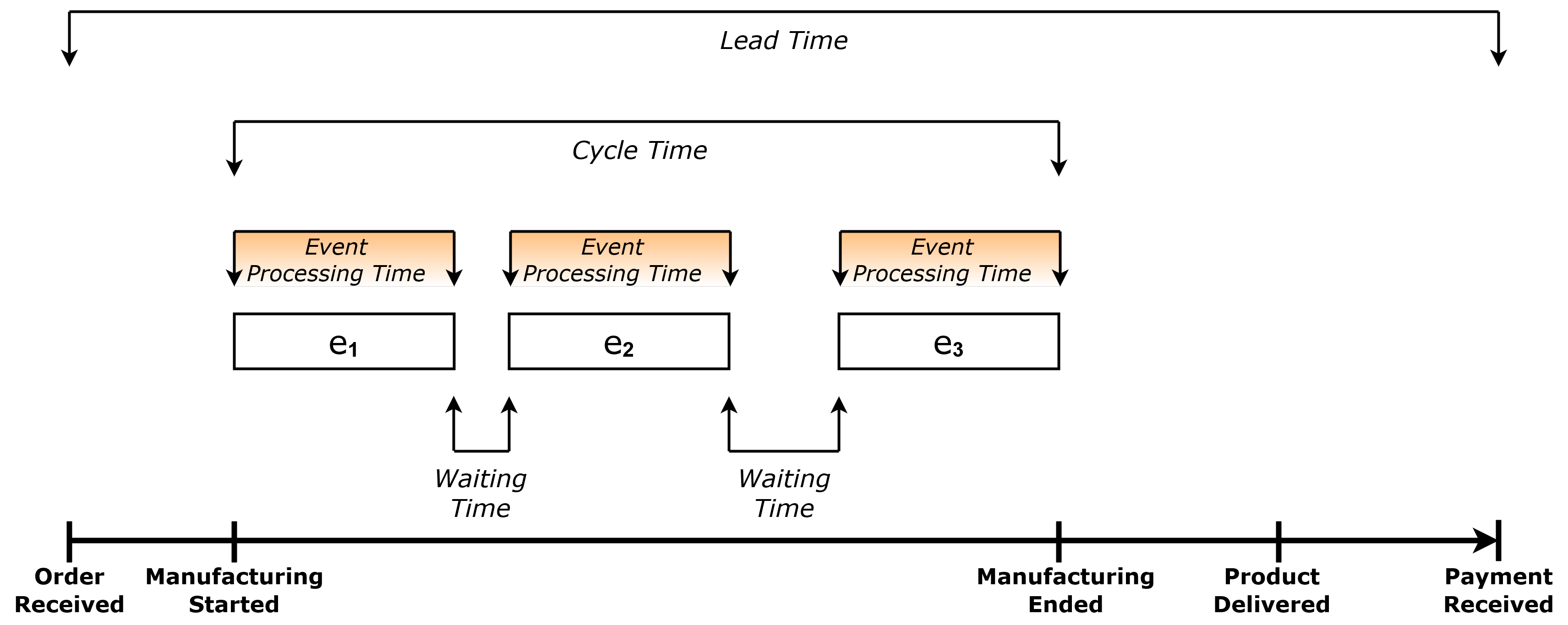}
\caption{Relationship of event processing time with other time concepts in the context of product manufacturing \cite{dumas2018introduction}}\label{Figure_Processing_Time_Diagram}
\end{figure}

Using the historical process event data, experts can now calculate the duration of each production activity, also referred to as event processing time, by determining the difference between the corresponding start and end timestamps. In a highly competitive industry, accurate estimation of event processing times is essential for effective planning. By accomplishing this, they can project the pure production time over the case to estimate the duration for the entire order. To illustrate the relationship of the event processing time to other prominent time concepts in the literature, refer to Figure \ref{Figure_Processing_Time_Diagram}. Event Processing time can be viewed as an element of the cycle time, which also encompasses waiting or idle time.

Upon identifying the target parameter of interest, probabilistic machine learning solutions should be employed to generate data-driven estimations and corresponding uncertainty information. This is supplemented with relevant explanation mechanisms, allowing users to comprehend the model uncertainty related to individual activity durations. The uncertainty-aware outputs are then utilized as input for decision augmentation scenarios or as input for the adopted optimization approach to generate production plans.

\section{Methodology} \label{Methodology}
\begin{figure}[h]%
\centering
\includegraphics[width = \textwidth]{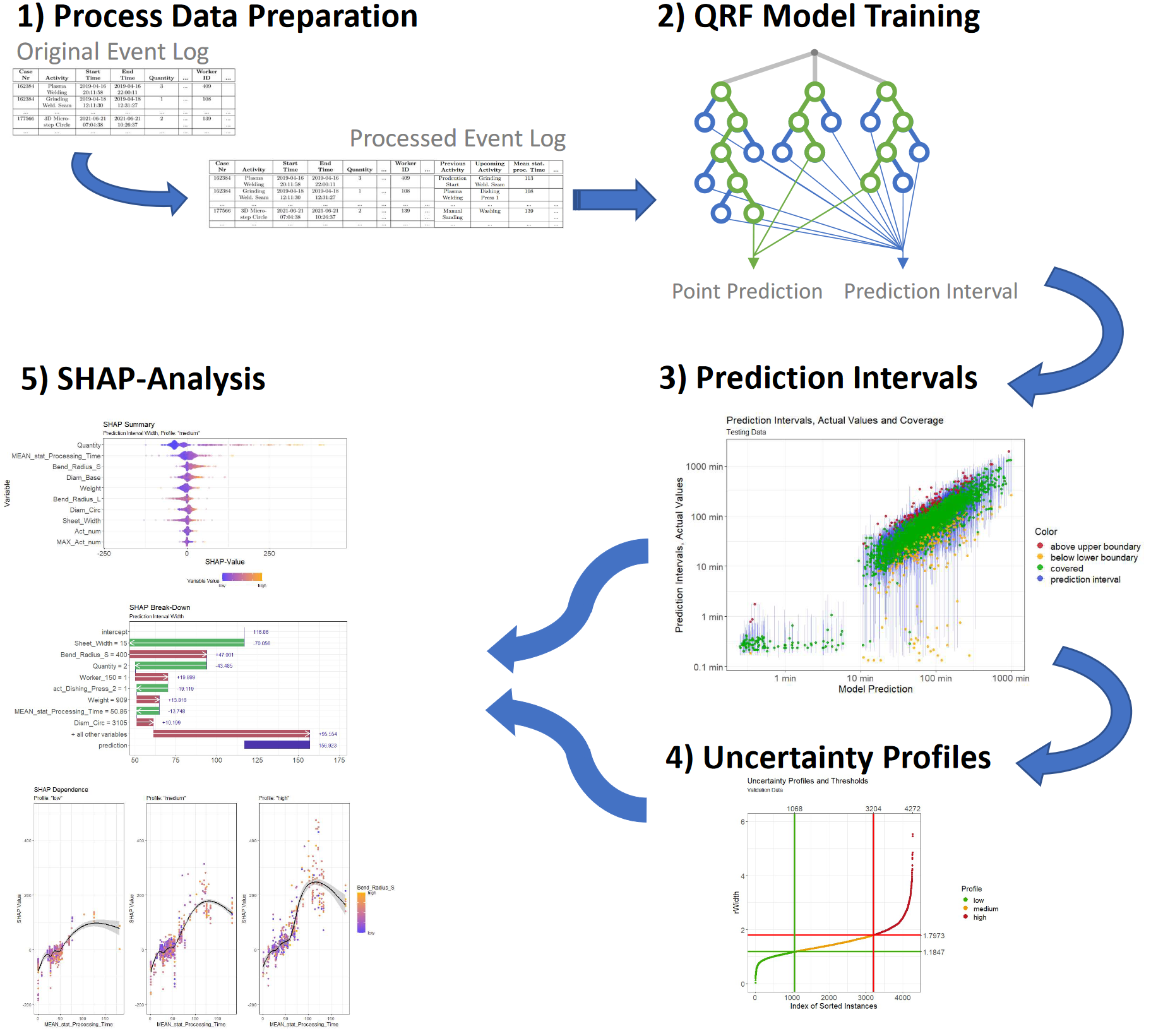}
\caption{Overview of the proposed uncertainty-aware approach}\label{Figure_Approach_Overview}
\end{figure}

The proposed approach presented in this study resides at the nexus of IS and AI, with a focus on addressing the inherent challenges of the OR problem. Our primary goal is to devise a sophisticated yet comprehensible machine learning-based solution by harnessing process event data supplied by the pertinent information systems (see Figure \ref{Figure_Approach_Overview}). In this context, it is crucial to acknowledge the OR implications, encompassing the necessity of quantifying uncertainty and ensuring model explainability while designing data-driven analytical solutions. This section provides an overview of the methodology, comprising steps involved in formal definition of process event data and its preparation for the supervised learning problem; utilization of QRF for uncertainty quantification; and implementation of the SHAP approach for elucidating model uncertainties.

\subsection{Process Data Preparation} \label{Process_Data_Preparation}
This section outlines the steps to create a tabular dataset from a process event log and to formulate the duration prediction for each activity, also referred to as event processing time, in the running traces as a supervised learning problem. To achieve this, it is essential to extract the predictors or input variables from the examined running traces and associate them with the corresponding target values. To enhance the comprehensibility of the underlying data structure, we begin by presenting notations and corresponding formal definitions of process data elements, such as events, event logs, traces, partial traces, and event duration, based on relevant literature \cite{polato2014data, van2016_process_mining, teinemaa2019_outcome_prediction}.

\begin{definition}[\textbf{Event}] An $event$ is a tuple $e=\left(a, c, t_{start}, t_{complete}, v_{1}, \ldots, v_{n}\right)$, where
\begin{itemize}
    \item $a \in \mathcal{A}$ is the corresponding process activity;
    \item $c \in \mathcal{C}$ is the case $i d$;
    \item $t_{start} \in \mathcal{T}_{start}$ is the start timestamp of the event (defined as seconds since $1 / 1 / 1970$ which is a Unix epoch time representation);
    \item $t_{complete} \in \mathcal{T}_{complete}$ is the completion timestamp of the event;
    \item $v_{1}, \ldots, v_{n}$ represents the list of event specific attributes, where $\forall  1 \leq i \leq n : v_{i} \in \mathcal{V}_{i}, \mathcal{V}_{i}$ denoting the domain of the $i^{th}$ attribute.
\end{itemize}
Consequently, $\mathcal{E}=\mathcal{A} \times \mathcal{C} \times \mathcal{T}_{start} \times \mathcal{T}_{complete} \times \mathcal{V}_{1} \times \cdots \times \mathcal{V}_{n}$ is defined as the universe of events. Moreover, we define the following project functions given the event $e \in \mathcal{E}$:
\begin{itemize}
    \item $p_a \colon \mathcal{E} \rightarrow \mathcal{A},\; p_{a}(e)=a$,
    \item $p_c \colon \mathcal{E} \rightarrow \mathcal{C},\; p_{c}(e)=c$, 
    \item $p_{{t_{start}}} \colon \mathcal{E} \rightarrow \mathcal{T}_{start},\; p_{{t_{start}}}(e)=t_{start}$,
    \item $p_{{t_{complete}}} \colon \mathcal{E} \rightarrow \mathcal{T}_{complete},\; p_{{t_{complete}}}(e)=t_{complete}$,
    \item $p_{v_i} \colon \mathcal{E} \rightarrow \mathcal{E}_i,\; p_{v_{i}}(e)=v_{i}, \forall 1 \leq i \leq n$
\end{itemize}
\end{definition}

\begin{definition}[\textbf{Traces and Event Log}] A $trace$ $\sigma \in \mathcal{E}^{*}$ is a finite sequence of events $\sigma_{c}=\langle e_{1}, e_{2}, \ldots, e_{{\lvert \sigma_{c} \rvert }}\rangle$, for which each $e_{i} \in \sigma$ occurs no more than once and $\forall e_i,e_j \in \sigma$, $p_{c}(e_{i})=p_{c}(e_{j}) \wedge$ $p_{\mathcal{T}_{S}}(e_{i}) \leqq p_{\mathcal{T}_{S}}\left(e_{j}\right)$, if $1 \leq i<j < {\lvert \sigma_{c} \rvert }$. The  \textit{event log $\mathcal{E}_C$} is defined as a set of completed traces, $\mathcal{E}_C=\left\{\sigma_{c} \mid c \in \mathcal{C}\right\}$.
\end{definition}

\begin{definition}[\textbf{Partial Traces}]
Two options are given for obtaining partial traces, depending on the predictive process monitoring use case. Defined over $\sigma$  the following $hd^{i}(\sigma_{c})$ and $tl^{i}(\sigma_{c})$ generate the \textit{prefixes} and \textit{suffixes} respectively as follows:
\begin{itemize}
\item selection operator (.): $\sigma_{c}(i)=\sigma_{i}, \forall 1 \leq i \leq n$;
\item $hd^{i}(\sigma_{c})=\langle e_{1}, e_{2}, \ldots, e_{\min (i, n)}\rangle$ for $i \in \left[1,{\lvert \sigma_{c} \rvert }\right] \subset \mathbb{N}$
\item $tl^{i}(\sigma_{c})=\langle e_{w}, e_{w+1}, \ldots, e_{n}\rangle$
where $w=\max (n-i+1,1)$;
\item ${\lvert \sigma \rvert }=n$ (i.e. the cardinality or length of the trace).
\end{itemize}
\end{definition}

We represent the set of partial traces generated using the $tl^{i}(\sigma_{c})$ function as $\gamma$. This can be subsequently utilized to formulate the tabular dataset for predicting the duration of each remaining event in the running trace. To further elucidate this point, it should be highlighted that our proposed approach to predictive process monitoring is executed prior to the initiation of the case. However, the problem and data structuring can be adapted to accommodate updates following each elapsed event in the running traces if necessary. Consequently, the application of the $tl^{i}(\sigma_{c})$ function may indeed be pertinent in generating partial traces for structuring training data. Various process performance indicators (PPIs) are frequently identified as targets of interest in predictive process monitoring studies. In addition to costs and quality, the proactive analysis of numerous time-related indicators is instrumental in augmenting the operational and strategic capabilities of organizations. We focus on the processing time of a specific event, which represents the duration of the corresponding activity, computed as the difference between the completion timestamp and the start timestamp of the event:

\begin{definition}[\textbf{Event Processing Time/Labeling}]
 Given a non-empty trace $\sigma \neq\langle\rangle \in \mathcal{E}^{*}$, a labeling function $resp: \mathcal{E} \rightarrow \mathcal{Y}$, also referred to as annotation function, maps an event $e \in \sigma$ to the corresponding value of its response variable $resp(e) \in \mathcal{Y}$. We define the event processing time as our response variable, calculated as follows:
\begin{equation}
\operatorname{\textit{resp}}(e)=p_{{t_{complete}}}(e)-p_{{t_{start}}}(e),
\end{equation}
with the domain of the defined response variable being $\mathcal{Y} \subset \mathbb{R}^+$.
\end{definition}

\begin{definition} [\textbf{Feature Extraction}]
The feature extraction function in this study is defined as a function $feat: \mathcal{E}^{*} \rightarrow \mathcal{X}^{*}$ which extracts the feature values from a given non-empty trace $\sigma \neq\langle\rangle \in \mathcal{E}^{*}$, with $\mathcal{X} \in \mathbb{R}^{dim}$ denoting the domain of the features and $dim$ being the input dimension. For a given trace $\sigma_{c}=\langle e_{1}, e_{2}, \ldots, e_{{\lvert \sigma_{c} \rvert }}\rangle$, the feature extraction function $feat$ generates a set of features $\left(x_{i, 1}, ..., x_{i, dim}\right)$ for each event $e_i$. n addition to case-specific and event-specific feature values, the feature extraction function enables the retrieval of intra-case-specific features, such as n-grams.
\end{definition} 

For a set of predictor variables $\mathbf{x} = \left(x_{1}, ..., x_{p}\right)$ and a response variable $y$, we define $D=\left\{\left(\mathbf{x_i}, y_i\right)\right\}_{i=1}^N$ as the dataset of associated $\left(\mathbf{x}, y\right)$ values, with N denoting total amount of observations. The dataset is split into three parts: $D=D_{train} \cup  D_{val} \cup D_{test}$, in order to use $D_{train}$ for training, $D_{val}$ for hyperparameter optimization and as a calibration set to derive uncertainty profiles (see Section \ref{Uncertainty_Profile_Construction_and_Evaluation}), and $D_{test}$ for evaluation, with $N_{train}$, $N_{val}$, $N_{test}$ being the respective amount of instances in each subset.

\subsection{Interval Prediction with Quantile Regression Forests} \label{Interval_Prediction_with_Quantile_Regression_Forests}

Random Forests (RF), as introduced by \cite{breiman2001random}, are a powerful machine learning technique that can provide a comprehensive understanding of the conditional distribution of response variables, not just their conditional means. This makes them useful in a variety of applications in operations research, where accurate predictions of uncertain outcomes are essential for effective decision-making. One important extension of RF is QRF, as introduced by \cite{Meinshausen}, which can be used to estimate conditional quantiles for high-dimensional predictor variables. This approach offers a non-parametric and accurate way of estimating prediction intervals, which provide valuable information about the dispersion of observations around the predicted value.

In accordance with \cite{Meinshausen}, let $\theta$ be a random parameter vector dictating tree growth within the RF, $T(\theta)$ the associated tree, $\mathcal{B}$ the space of a data point $X$ with dimensionality $p$ and $R_{\ell} \subseteq \mathcal{B}$ be a rectangular subspace for any leaf $\ell$ of every tree within the RF. Any $x \in \mathcal{B}$ can be allocated to exactly one leaf of any tree of the RF such that $x \in R_{\ell}$, thus denoting the specific leaf of the corresponding tree via $\ell(x, \theta)$.
First, the weight function $w_i(x, \theta)$ is defined for each observation $i$ and tree $T(\theta)$, given by

\begin{equation}
w_i(x, \theta)=\frac{1_{\left\{X_i \in R_{\ell(x, \theta)}\right\}}}{\#\left\{j: X_j \in R_{\ell(x, \theta)}\right\}}, 
\end{equation}

where $1_{\left\{X_i \in R_{\ell(x, \theta)}\right\}}$ is an indicator function that equals to 1 if the observation $i$ falls in the leaf node corresponding to $\ell(x, \theta)$, and $\#\left\{j: X_j \in R_{\ell(x, \theta)}\right\}$ is the number of observations that fall in that same leaf node.

Next, the weights are averaged over multiple trees to obtain the final weight function $w_i(x)$:

\begin{equation}
w_i(x)=k^{-1} \sum_{t=1}^k w_i\left(x, \theta_t\right),
\end{equation}

where $k$ is the number of trees and $\theta_t$ represents the $t$-th tree.

Finally, the estimated conditional quantile function $\widehat{F}(y \mid X=x)$ is calculated as a weighted sum of the indicator function $1_{\left\{Y_i \leq y\right\}}$ for each observation $i$, where $Y_i$ is the response variable:

\begin{equation}
\widehat{F}(y \mid X=x)=\sum_{i=1}^n w_i(x) 1_{\left\{Y_i \leq y\right\}}.
\end{equation}

Using this estimated conditional quantile function, the quantile function $Q_\alpha(x)$ for a given quantile level $\alpha$ can be obtained as

\begin{equation}
Q_\alpha(x)=\inf \{y: \widehat{F}(y \mid X=x) \geq \alpha\}.
\end{equation}

Based on the quantile function $Q_\alpha(x)$, prediction intervals of specific levels can be derived using

\begin{equation}\label{Equation_Prediction_Interval}
I_{1-2\alpha}(x)=\left[Q_{\alpha}(x), Q_{1-\alpha}(x)\right],
\end{equation}

where $1-2\alpha$ denotes the level of the prediction interval. For example, given $\alpha = 5\%$, equation (\ref{Equation_Prediction_Interval}) yields the $90\%$-prediction interval $I_{90\%}(x)=\left[Q_{5\%}(x), Q_{95\%}(x)\right]$.

For the prediction of the conditional mean, as yielded by regular RF, QRF allow the utilization of

\begin{equation}
\widehat{F}(y \mid X=x)=\sum_{i=1}^n w_i(x)Y_i.
\end{equation}

In operations research, prediction intervals can be used to quantify uncertainty in various decision-making contexts. For example, they can be used to estimate the probability of an outcome exceeding a certain threshold or to evaluate the risk associated with different decision options. This information is critical for making informed decisions and optimizing operational efficiency. This is because standard prediction models that return a single point estimate for each new instance may not capture the full complexity of the underlying data, which can lead to inaccurate predictions and suboptimal decision-making.

\subsection{Explanations with SHAP} \label{Explanations_with_SHAP}

In accordance with \cite{LIME}, a prediction model $f$ for can be approximated for a locality by an interpretable explanation model $g$, employing simplified inputs $x^{\prime}$ that fulfill $x = h_x(x^{\prime})$, with $h_x$ being a mapping function specific to the original input $x$. For $z^{\prime} \approx x^{\prime}$, local methods approximate $f(h_x(z^{\prime}))$ via $g(z^{\prime})$. Additive feature attribution methods employ the following explanation model:

\begin{equation}\label{Equation_Additive_Feature_Attribution}
g\left(z^{\prime}\right)=\phi_0+\sum_{i=1}^M \phi_i z_i^{\prime},
\end{equation}

with $z^{\prime} \in\{0,1\}^M, M$ being the number of simplified input features and $\phi_i \in \mathbb{R}$ being the feature effect. The sum of $\phi_0$ and feature effects allows for the approximation of $f(x)$. In order to derive $\phi_i$, the LIME approach minimizes the objective $\xi$-function through penalized linear regression:

\begin{equation}\label{LIME}
\xi=\underset{g \in \mathcal{G}}{\arg \min } L\left(f, g, \pi_{x^{\prime}}\right)+\Omega(g),
\end{equation}

where $L$ is squared loss, $\pi_{x^{\prime}}$ is a local kernel weighting simplified inputs and $\Omega$ a penalty function for the complexity of $g$. Shapley Regression Values, as presented in \cite{Shapley_Regression_Values}, measure changes in the expected prediction depending on the inclusion or exclusion of specific features. Given a subset $S \subseteq F$ with $F$ consisting of all features and a specific feature $i \in F$, a model $f_{S \cup\{i\}}$ containing the feature $i$ and a model $f_S$ excluding the feature $i$ are trained, with $x_{S \cup\{i\}}$ and $x_S$ being the corresponding input vectors of feature values. The predictions of the two models are compared via $\left[f_{S \cup\{i\}}\left(x_{S \cup\{i\}}\right)-f_S\left(x_S\right)\right]$ and by iterating over all possible subsets $S \subseteq F\backslash\{i\}$ of features, the Shapley values are calculated as the weighted mean over all possible differences:

\begin{equation}
\phi_i=\sum_{S \subseteq F \backslash\{i\}} \frac{\lvert S\rvert !(\lvert F\rvert-\lvert S\rvert-1) !}{\lvert F\rvert !}\left[f_{S \cup\{i\}}\left(x_{S \cup\{i\}}\right)-f_S\left(x_S\right)\right]
\end{equation}

In accoradance with \cite{SHAP_Lundberg}, the consistency, local accuracy and missingness properties are satisfied by

\begin{equation}\label{Equation_Shapley_Values}
\phi_i(f, x)=\sum_{z^{\prime} \subseteq x^{\prime}} \frac{\lvert z^{\prime}\rvert !\left(M-\lvert z^{\prime}\rvert-1\right) !}{M !}\left[f_x\left(z^{\prime}\right)-f_x\left(z^{\prime} \backslash i\right)\right],
\end{equation}

with $\lvert z^{\prime}\rvert$ being the number of non-zero entries in $z^{\prime}$, and $z^{\prime} \subseteq x^{\prime}$ representing all $z^{\prime}$ vectors with non-zero entries being a subset of the non-zero entries in $x^{\prime}$. In this context, SHAP values are proposed by \cite{SHAP_Lundberg} as a solution to equation (\ref{Equation_Shapley_Values}), with $f_x$ as the conditional expectation function of the original model, fulfilling $f_x(z^{\prime}) = f(h_x(z^{\prime})) = E\left[f(z)\vert z_S \right]$, with $S$ being the set of non-zero indexes in $z^{\prime}$. Furthermore, KernelSHAP is proposed as an approximation to the computationally demanding calculation of exact SHAP values, with the following parameters satisfying equation (\ref{LIME}):

\begin{equation}
\begin{aligned}
\Omega(g) & =0, \\
\pi_{x^{\prime}}\left(z^{\prime}\right) & =\frac{(M-1)}{\left(M \text { choose }\lvert z^{\prime}\rvert \right)\lvert z^{\prime}\rvert \left(M-\lvert z^{\prime}\rvert \right)}, \\
L\left(f, g, \pi_{x^{\prime}}\right) & =\sum_{z^{\prime} \in Z}\left[f\left(h_x^{-1}\left(z^{\prime}\right)\right)-g\left(z^{\prime}\right)\right]^2 \pi_{x^{\prime}}\left(z^{\prime}\right),
\end{aligned}
\end{equation}

with $\lvert z^{\prime}\rvert$ being the number of non-zero elements in $z^{\prime}$, employing linear LIME while assuming model linearity and feature independence for the corresponding locality. For the implementation of KernelSHAP, the R library \textit{"kernelshap"} was used.

\section{Experiment Settings} \label{Experiment_Settings}
This section gives a general overview of the underlying dataset, the tools employed for the implementation of our proposed approach, hyperparameter optimization for the QRF model as well as a description of utilized model evaluation metrics.
\subsection{Dataset Overview and Tools} \label{Data_Set_Overview_and_Tools}
Table \ref{Data_Set_Information} presents a general statistic for the underlying event log. The original dataset consists of 12,077 cases with a total 55,846 events, pertaining to a total of 30 unique activities. A mean processing time of 98.22 minutes with a standard deviation of 143.96 minutes and a mean trace length of 4.624 events per case has been observed.
\begin{table}[h]
\begin{center}
\begin{minipage}{330pt}
\caption{Data Set Information}\label{Data_Set_Information}
\begin{tabular*}{330pt}{@{\extracolsep{\fill}}l||r|r|r||r@{\extracolsep{\fill}}}
\toprule
                    & Train  & Validation    & Test   & Complete Data Set        \\
\hline
\hline
Number of events    & 47966     & 4272          & 3608      & 55846       \\
\hline
Number of cases     & 10265     & 906           & 906       & 12077       \\
\hline
Unique activities   & 30        & 26            & 27        & 30       \\
\hline
Mean processing     & 98.36     & 100.80        & 93.28     & 98.22 \\
time (min)          & & & & \\
\hline
Std. deviation of   & 145.05    & 154.54        & 113.01     & 143.96 \\
processing time (min)          & & & & \\
\hline
Mean trace          & 4.67      & 4.72          &  3.98      & 4.624 \\
length              & & & & \\
\botrule
\end{tabular*}
\end{minipage}
\end{center}
\end{table}
For model training, the dataset was split into training, validation and test sets. Since the data is comprised of process steps belonging to certain production cases, training, validation and test data were split in a ratio of 85 : 7.5 : 7.5 based on the case identifier. This guarantees that process steps belonging to the same production case are not split across various datasets.  To conform with the usage scenario, the data was split in chronological order, thus assigning the oldest cases to the training set and the latest cases to the validation and test sets. This results in a distribution of instances that does not fully reflect the split ratio: the training data consist of 47,966, the validation data consists of 4,272 and the test data consists of 3,608 instances, respectively representing 85.89\%, 7.65\% and 6.46\% of available data. The corresponding information for these data subsets can be found in table \ref{Data_Set_Information} as well.

The implementation of the QRF was carried out in \textit{R}, mainly via the libraries \textit{"tidymodels"} and \textit{"ranger"}. For data cleaning, processing and feature engineering, the \textit{"tidyverse"} library-collection was used, predominantly \textit{"dplyr"}. For the calculation of SHAP-values, \textit{"kernelshap"} in conjunction with \textit{"doParallel"} was employed, and for visualization, \textit{"ggplot2"}, \textit{"ggstatsplot"} and \textit{"ggbeeswarm"} were used predominantly.
\subsection{Model Training and Hyperparameter Optimization} \label{Model_Training_and_Hyperparameter_Optimization}
In order to increase the performance of the QRF model with regards to point prediction, hyperparameter optimization was performed for the following model specifications: \textit{mtry} represents the number of predictor variables that are sampled randomly for splits in the tree models, \textit{trees} determines the total number of tree models of the ensemble and \textit{min\_n} marks the minimum number of observations in a node which allows further splitting. Root Mean Squared Error (RMSE) was chosen as the deciding metric for choosing the optimal model.

\begin{table}[h]
\begin{center}
\begin{minipage}{300pt}
\caption{Hyperparameter Optimization for Quantile Regression Forest}\label{Table_Hyperparameter_Optimization}
\begin{tabular}{@{}lrrr@{}}
\toprule
Hyperparameter      & Search Space      & Search Grid Values            & Selected Value \\
\midrule
mtry                & [40, 90]          & (40, 50, 60, 70, 80, 90)      & 70 \\
trees               & [50, 100]         & (50, 60, 70, 80, 90, 100)     & 100 \\
min\_n              & [2, 32]           & (2, 8, 14, 20, 26, 32)        & 20 \\
\botrule
\end{tabular}
\end{minipage}
\end{center}
\end{table}

Table \ref{Table_Hyperparameter_Optimization} describes the ranges for each hyperparameter, from which 6 equidistant values were sampled, resulting in $6^3 = 216$ combinations. The best performing model with regards to the RMSE was trained with \textit{mtry} $= 70$, \textit{trees} $= 100$, \textit{min\_n} $= 20$ and is evaluated in Section \ref{Model_Evaluation} in detail.
\subsection{Evaluation Metrics} \label{Evaluation_Metrics}
In order to evaluate the model performance of the adopted machine learning approach, we scrutinize two aspects of the predictive strength. Firstly, concerning point predictions, the assessment of the predictive quality for regression models can rely on a range of metrics that endeavor to capture the prediction error in relation to the underlying ground truth value. In this study, we employ RMSE and Mean Absolute Error (MAE) as the chosen metrics. The RMSE is derived by computing the squared differences between the predicted and actual values for the given dataset, followed by calculating the square root of the average:

\begin{equation}
RMSE=\sqrt{\frac{1}{N}\sum_{i=1}^{N}\left(y_{i}-\hat{y_{i}}\right)^{2}}
\end{equation}

As posited by \cite{willmott2005advantages}, the RMSE is considerably influenced by the distribution of error magnitudes, which may render it misleading for assessing model quality in specific instances. Consequently, the authors propose MAE as a more comprehensible metric. The MAE is calculated by determining the average of the absolute differences between the predicted and actual values for the given dataset:

\begin{equation}
MAE = \frac{1}{N} \sum_{i=1}^{N} {{\lvert y_{i}-\hat{y_{i}} \rvert }}
\end{equation}

Following the suggestion in \cite{chai2014root}, where the RMSE was demonstrated to be a more reliable evaluation metric in cases with Gaussian error distribution, both metrics will be employed in this study to capture model performance.

The second evaluation dimension pertains to the quality of prediction intervals, which can be assessed based on metrics that capture the relationship between the prediction interval and the actual value. We propose Prediction Interval Coverage Probability (PICP) and Mean Prediction Interval Width (MPIW) as in \cite{PICP_MPIW_1}, along with Mean Relative Prediction Interval Width (MRPIW) as in \cite{JORGENSEN200479, rMPIW}. The PICP represents the proportion of target values covered by the prediction intervals and is calculated as the ratio between the number of instances whose target values were covered by the corresponding prediction interval and the total number of instances:

\begin{equation}
\mathrm{PICP}=\frac{1}{N} \sum_{i=1}^N c_i, \quad c_i= \begin{cases}1, & y_i \in\left[L_i, U_i\right] \\ 0, & y_i \notin\left[L_i, U_i\right]\end{cases}
\end{equation}

The MPIW further aids in assessing the quality of the generated intervals and is calculated by averaging the distances between the upper and lower boundaries of the prediction intervals:

\begin{equation}
MPIW = \frac{1}{N} \sum_{i=1}^{N} {{( U_i- L_i )}}
\end{equation}

Since the MPIW seeks to capture model uncertainty through interval width in absolute terms, \cite{JORGENSEN200479, rMPIW} advocate for a metric that encompasses the relative relationship between the predicted ranges and the estimated outcome. Specifically, the proposed normalization via point prediction computes a relative interval width (rWidth) using the quotients between prediction interval widths and corresponding point estimations. The MRPIW is the average rWidth of the evaluated instances:

\begin{equation}
MRPIW = \frac{1}{N} \sum_{i=1}^{N} {{ rWidth_i }}, \quad rWidth_i = \frac{(U_i- L_i)}{\hat{y_{i}}} 
\end{equation}

In this context, \cite{JORGENSEN200479} contend that model uncertainty decreases with lower MRPIW values for datasets with the same PICP. For the evaluation of model uncertainty, all three of the aforementioned metrics will be employed.

\section{Results} \label{Results}
This section analyzes the results of the proposed approach, utilizing the QRF model obtained from the hyperparameter optimization presented in Section \ref{Model_Training_and_Hyperparameter_Optimization}. First, the QRF model is evaluated in general based on the metrics outlined in Section \ref{Evaluation_Metrics}. Next, the established uncertainty profiles are presented, and the model performance for each profile is analyzed. The section concludes by examining SHAP-based methods and comparing the influence of feature values on point predictions and prediction intervals, both generally and across uncertainty profiles.
\subsection{Model and Uncertainty Evaluation} \label{Model_Evaluation}
This subsection presents the outcomes of the model evaluation. First, the model's predictive strength for point prediction is assessed, followed by an evaluation of its ability to capture uncertainty through prediction intervals. Lastly, the model's performance for a selected set of process activities within the dataset is examined. To adhere to the intended usage scenario, uncertainty profiles must be generated on the validation data to derive reliable MRPIW thresholds. Instances from the test data are then assigned to the appropriate uncertainty profile based on these thresholds. Consequently, the evaluation of the model's performance on the validation data is also incorporated in the analysis process.

Table \ref{Table_Model_Evaluation_Point_Prediction} presents the results of the model performance evaluation for the validation and test datasets concerning point predictions, as evaluated by the MAE and RMSE metrics.
\begin{table}[h]
\begin{center}
\begin{minipage}{200pt}
\caption{Evaluation Metrics for the Quantile Regression Forest (Point Prediction)}\label{Table_Model_Evaluation_Point_Prediction}
\begin{tabular*}{200pt}{@{\extracolsep{\fill}}lrr@{\extracolsep{\fill}}}
\toprule
Data Set            & MAE       & RMSE \\
\midrule
Validation Data     & 35.045    & 81.66 \\
Test Data        & 35.51     & 66.88 \\
\botrule
\end{tabular*}
\end{minipage}
\end{center}
\end{table}

Although the difference in MAE values between validation and test data is slim, a greater difference between RMSE can be observed. This difference can arguably be attributed to the chronological data splitting (see Section \ref{Model_Training_and_Hyperparameter_Optimization}), for example due to prominence of certain production orders and their specifications within the time frames captured in the validation and test datasets.

Table \ref{Table_Model_Evaluation_Uncertainty} displays the results of the model's performance evaluation regarding the prediction intervals produced by the QRF model. As outlined in Section \ref{Interval_Prediction_with_Quantile_Regression_Forests}, the proposed approach generated a PICP of approximately $\sim90\%$ for both the validation and test datasets, meeting the expectations of the process owners. Comparing the datasets, the MRPIW for the test data (1.492) was slightly lower than that for the validation data (1.561), indicating a lower level of uncertainty in the model's predictions for the test dataset. On the other hand, the MPIW was slightly larger for the test data (154.69) than for the validation data (152.94). These findings suggest that the model was able to capture the uncertainty inherent in the data and provide prediction intervals that were reliable for both datasets. Overall, the results demonstrate that the model was capable of accurately predicting the duration of process events while providing a measure of uncertainty in the form of prediction intervals.
\begin{table}[h]
\begin{center}
\begin{minipage}{200pt}
\caption{Evaluation Metrics for the Quantile Regression Forest (Uncertainty)}\label{Table_Model_Evaluation_Uncertainty}
\begin{tabular*}{200pt}{@{\extracolsep{\fill}}lrrr@{\extracolsep{\fill}}}
\toprule
Data Set            & PICP      & MPIW      & MRPIW \\
\midrule
Validation Data     & 0.893     & 152.94    & 1.561 \\
Test Data        & 0.895     & 154.69    & 1.492 \\
\botrule
\end{tabular*}
\end{minipage}
\end{center}
\end{table}

\begin{figure}[h]%
\centering
\includegraphics[width = 0.9\textwidth]{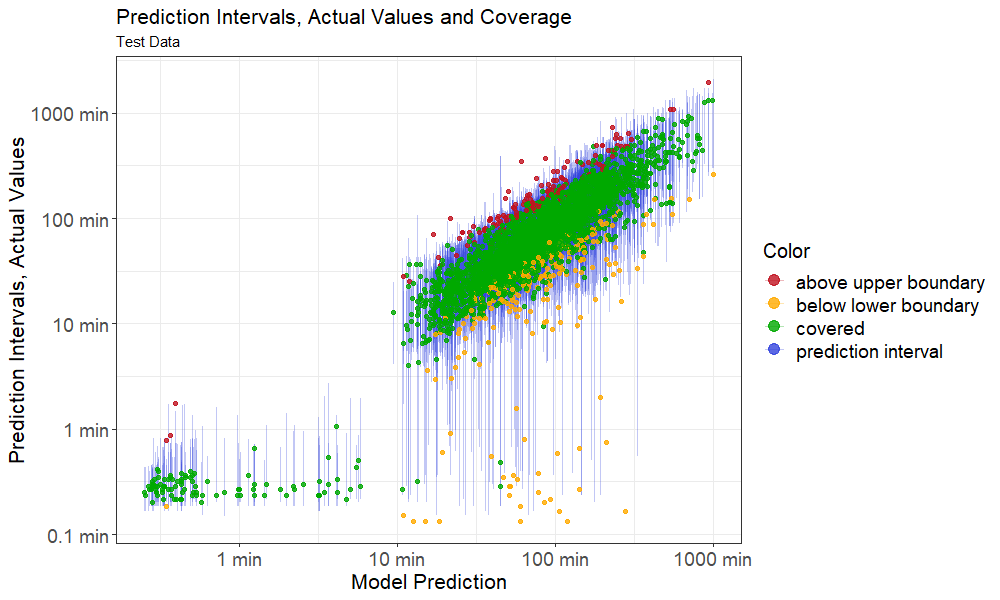}
\caption{Visualization of model predictions, actual values, prediction intervals and coverage for the test data set. The points represent the relationship between the model predictions on the x-axis and the actual values on the y-axis. Corresponding prediction intervals are depicted as blue vertical lines, spanning from the value of the upper boundary to the value of the lower boundary. The color of the points indicate if the actual value was captured by the prediction interval (green) or not (orange for actual values below lower boundary, red for actual values above upper boundary). A logarithmic scale was used to allow a clearer depiction of values below the lower boundary.}\label{Figure_Prediction_Intervals_Coverage}
\end{figure}

Figure \ref{Figure_Prediction_Intervals_Coverage} provides a visual representation of the relationship between the actual values, point predictions, and prediction intervals for the test dataset. The majority of the values lie between 10 and 250 minutes, but some faulty predictions are due to the actual value being below the lower boundary of the prediction interval. Out of the total values, $10.5\%$ were not covered by the prediction intervals, with $3.64\%$ of these values being above the upper boundary and $6.86\%$ being below the lower boundary. Empirically, if the prediction interval does not capture the actual value, it is found that the actual value exceeds the upper boundary in $34.7\%$  of the cases and falls below the lower boundary in $65.3\%$  of the cases.

\begin{figure}[h]%
\centering
\includegraphics[width = 0.9\textwidth]{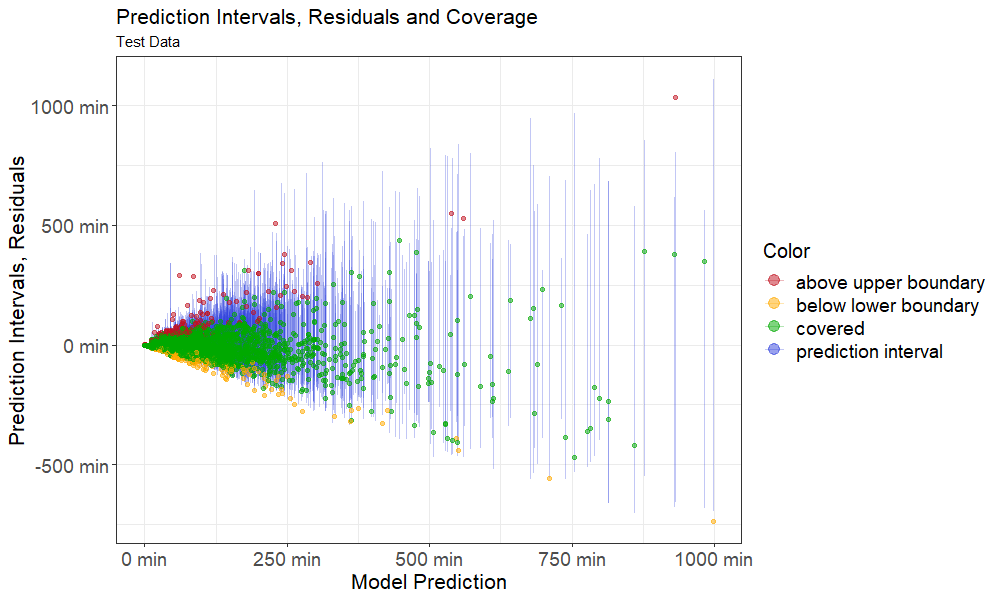}
\caption{Visualization of model predictions, residuals, prediction intervals and coverage for the test data set. The points represent the relationship between the model predictions on the x-axis and residual values on the y-axis. Corresponding prediction intervals are depicted as blue vertical lines, spanning from the value of the upper boundary to the value of the lower boundary reduced by the model prediction. The color of the points indicate if the actual value was captured by the prediction interval (green) or not (orange for actual values below lower boundary, red for actual values above upper boundary).}\label{Figure_Prediction_Intervals_Residuals}
\end{figure}

Figure \ref{Figure_Prediction_Intervals_Residuals} depicts a graphical representation of the residuals that exist between the actual values and the model predictions in the test dataset. This visualization corroborates the findings outlined in Figure \ref{Figure_Prediction_Intervals_Coverage}, providing evidence that a substantial proportion of inaccurate predictions are due to the inability of the lower boundary to encapsulate the actual values.

As presented in Section \ref{Process_Data_Preparation}, the implementation of a particular activity type serves as a crucial determinant for predicting the duration of its corresponding process step. In order to assess the model performance in relation to the process activity, three distinct activities were chosen, and the predictive strength of the QRF model was appraised utilizing a subset of the test data corresponding to these activities. The activities selected for this analysis are \textit{"Edge\_Bead," "Dishing\_Press\_1," and "Plasma\_Welding"} which respectively represent the processing of material on a specific dishing press, the refinement of the material with respect to its edge bead, and the processing of material via plasma welding. 

Table \ref{Table_Model_Evaluation_Point_Prediction_per_Activity} illustrates the outcomes of this evaluation, revealing substantial discrepancies in model performance among these activities. Regarding point predictions, "Edge\_Bead" exhibits the lowest MAE and RMSE values among the trio of selected activities, succeeded by "Dishing\_Press\_1," and subsequently "Plasma\_Welding." These metrics are coherent when considered within the context of the mean duration and standard deviation of these activities for the entire dataset: "Edge\_Bead," "Dishing\_Press\_1," and "Plasma\_Welding" display mean durations of 16.54 minutes, 90.34 minutes, and 134.66 minutes, respectively, along with standard deviations of 15.03 minutes, 90.3 minutes, and 144.73 minutes, respectively. 
\begin{table}[h]
\begin{center}
\begin{minipage}{200pt}
\caption{Evaluation Metrics for the Quantile Regression Forest for Selected Activities (Point Prediction)}\label{Table_Model_Evaluation_Point_Prediction_per_Activity}
\begin{tabular*}{200pt}{@{\extracolsep{\fill}}lrr@{\extracolsep{\fill}}}
\toprule%
Data Set                        & MAE       & RMSE      \\
\midrule  
Test Data                    &           &           \\
\hline
Activity: "Edge\_Bead"          & 5.64      & 7.61      \\
Activity: "Dishing\_Press\_1"   & 35.24     & 61.97     \\
Activity: "Plasma\_Welding"     & 37.99     & 59.47     \\
\botrule
\end{tabular*}
\end{minipage}
\end{center}
\end{table}

Upon examining the uncertainty measures, notable disparities among the selected activities become apparent (see Table \ref{Table_Model_Evaluation_Uncertainty_per_Activity}). In terms of prediction interval coverage, "Edge\_Bead" takes the lead with a PICP of 0.892, followed by "Dishing\_Press\_1" at 0.885, and "Plasma\_Welding" at 0.840. Nevertheless, the inferior coverage for "Plasma\_Welding" is accompanied by reduced model uncertainty in comparison to the other two activities: "Plasma\_Welding" manifests the lowest model uncertainty with an MPIW of 145.143 and an MRPIW of 1.379, succeeded by "Dishing\_Press\_1" (MPIW: 135.240, MRPIW: 1.638) and "Edge\_Bead" (MPIW: 32.665, MRPIW: 1.734).
\begin{table}[h]
\begin{center}
\begin{minipage}{240pt}
\caption{Evaluation Metrics for the Quantile Regression Forest for Selected Activities (Uncertainty)}\label{Table_Model_Evaluation_Uncertainty_per_Activity}
\begin{tabular*}{240pt}{@{\extracolsep{\fill}}lrrr@{\extracolsep{\fill}}}
\toprule%
Data Set                           & PICP      & MPIW    & MRPIW \\
\midrule
Test Data                       &           &           &  \\
\hline
Activity: "Edge\_Bead"             & 0.892     & 32.665    & 1.734 \\
Activity: "Dishing\_Press\_1"      & 0.885     & 135.240   & 1.638 \\
Activity: "Plasma\_Welding"        & 0.840     & 145.143   & 1.379 \\
\botrule
\end{tabular*}
\end{minipage}
\end{center}
\end{table}

\subsection{Uncertainty Profile Construction and Evaluation} \label{Uncertainty_Profile_Construction_and_Evaluation}
\begin{figure}[h]%
\centering
\includegraphics[width = 0.8\textwidth]{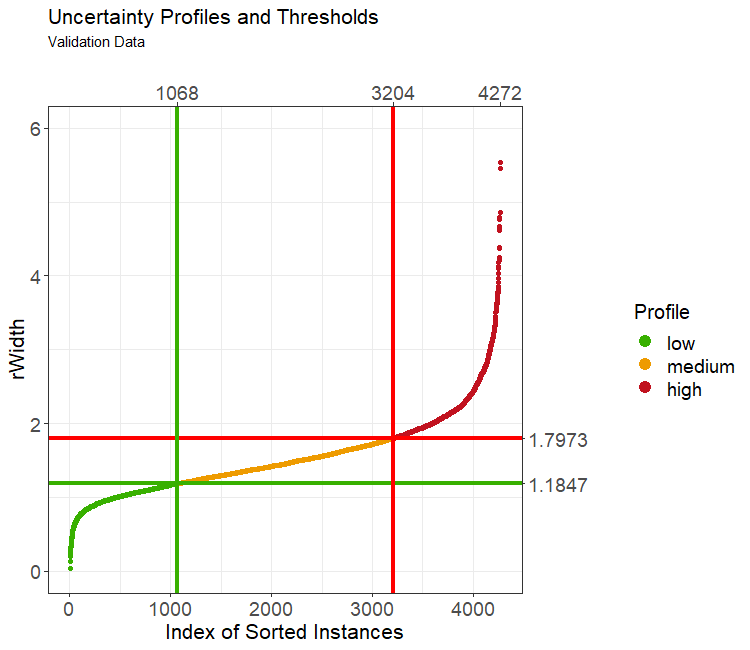}
\caption{Visualization of uncertainty profile thresholds for validation data. First, data set was sorted in ascending order of rWidth values and a numerical index was introduced to depict the order. Each point in the plot depicts its index on the x-axis and the corresponding rWidth value on the y-axis. The green and red vertical line divide points respectively at the 25th and 75th percentile. The green and red horizontal lines depict the corresponding rWidth values, which respectively separate the "low" from the "medium" (rWidth = 1.1847) and the "medium" from the "high" profile (rWidth = 1.7973). The points are colored to represent their affiliation with the corresponding profile: green for "low" , orange for "medium" and red for "high".}\label{Figure_Uncertainty_Profiles_Validation}
\end{figure}

Process experts have expressed interest in categorizing model uncertainties into distinct profiles, specifically low, medium, and high. By doing so, they aim to identify which model predictions require cautious interpretation and subsequent adjustments. Consequently, in addition to providing prediction intervals as uncertainty measures, we assign the model predictions to so-called uncertainty profiles. Initially, percentile-based profiling is considered in terms of prediction intervals. However, this approach may yield inaccurate outcomes, as the prediction intervals are significantly influenced by the actual output values. Activities with longer durations tend to exhibit larger prediction intervals, whereas those with shorter durations demonstrate the opposite trend. To address this issue, we employ the relative width intervals introduced in Section \ref{Evaluation_Metrics} for generating uncertainty profiles. This method allows for a more accurate representation of the uncertainty associated with each prediction, regardless of the activity duration. By categorizing model uncertainties using relative width intervals, process experts can better understand the level of confidence associated with each prediction and make informed decisions regarding potential adjustments (see Figure \ref{Figure_Uncertainty_Profiles_Validation}).

The construction of uncertainty profiles involved scoring the validation dataset using the fitted QRF model and calculating the rWidth for each model prediction. Instances were then sorted in ascending order of rWidth values, and the 25th and 75th percentile thresholds were used to define the uncertainty profiles. Values below the 25th percentile threshold were classified as "low" profile, with rWidth values below 1.1847, while values above the 75th percentile threshold were classified as "high" profile, with rWidth values above 1.7973. The remaining values were assigned to the "medium" profile. Figure \ref{Figure_Uncertainty_Profiles_Validation} visualizes the results, with the vertical lines indicating the 25th and 75th percentile thresholds and the horizontal lines representing the rWidth values corresponding to these thresholds.

Table \ref{Table_Model_Evaluation_per_Profile_Uncertainty_Validation} presents the PICP, MPIW, and MRPIW for each profile of the validation data to evaluate model uncertainty. For the PICP metric, the "high" profile has the highest value (0.909), followed by the "medium" profile (0.891), and the "low" profile (0.882). This trend suggests that the QRF model has better coverage for predictions with higher uncertainty. For the MPIW metric, the "high" profile exhibits the widest prediction intervals (210.0), followed by the "medium" profile (146.0), and the "low" profile (109.0). This result is expected as higher uncertainty profiles are associated with larger prediction intervals. Regarding the MRPIW metric, the "high" profile has the highest value (2.32), followed by the "medium" profile (1.47), and the "low" profile (0.985). This trend indicates that the relative width of the prediction intervals is larger for predictions with higher uncertainty. In summary, Table \ref{Table_Model_Evaluation_per_Profile_Uncertainty_Validation} demonstrates that the QRF model has better coverage for predictions with higher uncertainty, but also exhibits larger prediction intervals and relative widths for these predictions. This result highlights the trade-off between model coverage and uncertainty, suggesting that the model performs better in terms of coverage for the "high" uncertainty profile at the cost of increased uncertainty.
\begin{table}[h]
\begin{center}
\begin{minipage}{240pt}
\caption{Evaluation Metrics for the Quantile Regression Forest for Uncertainty Profiles of Validation Data  (Uncertainty)}\label{Table_Model_Evaluation_per_Profile_Uncertainty_Validation}
\begin{tabular*}{240pt}{@{\extracolsep{\fill}}lrrr@{\extracolsep{\fill}}}
\toprule%
Data Set            & PICP      & MPIW      & MRPIW \\
\midrule  
Validation Data     &           &           &  \\
\hline
Profile: "low"        & 0.882     & 109.0     & 0.985 \\
Profile: "medium"     & 0.891     & 146.0     & 1.47 \\
Profile: "high"       & 0.909     & 210.0     & 2.32 \\
\botrule
\end{tabular*}
\end{minipage}
\end{center}
\end{table}

\begin{figure}[h]%
\centering
\includegraphics[width = 0.8\textwidth]{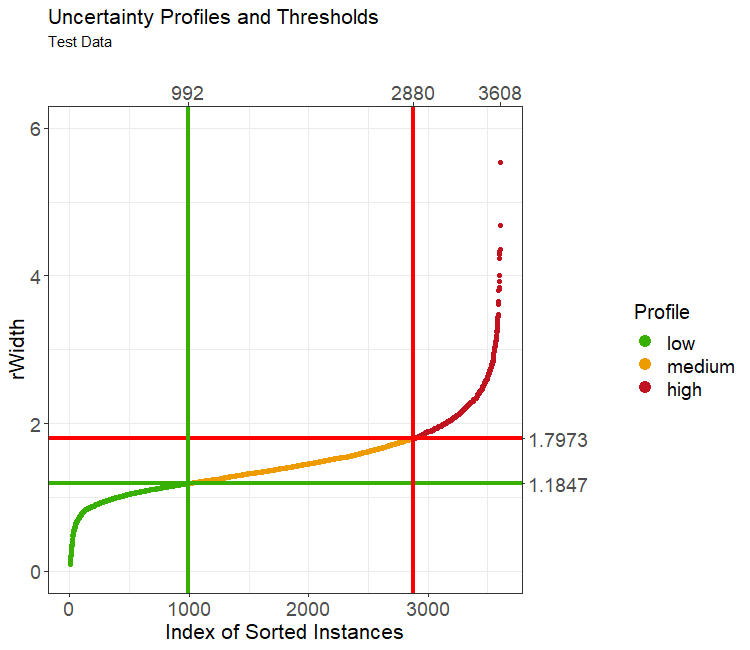}
\caption{Visualization of uncertainty profile thresholds for test data. First, data set was sorted in ascending order of rWidth values and a numerical index was introduced to depict the order. Each point in the plot depicts its index on the x-axis and the corresponding rWidth value on the y-axis. The green and red horizontal lines depict the rWidth-thresholds based on the results from the validation data, which respectively separate the "low" from the "medium" (rWidth = 1.1847) and the "medium" from the "high" profile (rWidth = 1.7973). The green and red vertical line depict the index at which the separation between profiles takes place. The points are colored to represent their affiliation with the corresponding profile: green for "low" , orange for "medium" and red for "high".}\label{Figure_Uncertainty_Profiles_Test}
\end{figure}

Similar to Figure \ref{Figure_Uncertainty_Profiles_Validation}, Figure \ref{Figure_Uncertainty_Profiles_Test} displays the allocation of uncertainty profiles to scored instances from the test dataset. Since the profile thresholds were calibrated on the rWidth values from the validation data, applying these thresholds does not divide the test data into the same ratio as the validation data: out of 3,608 instances, 992 instances ($27.5\%$) belong to the "low" profile, 1,888 instances ($52.3\%$) to the "medium" profile, and 728 instances ($20.2\%$) to the "high" profile.

Table \ref{Table_Model_Evaluation_per_Profile_Uncertainty_Test} presents the PICP, MPIW, and MRPIW for each profile of the test data to evaluate model uncertainty, following similar trends as documented in Table \ref{Table_Model_Evaluation_per_Profile_Uncertainty_Validation} for validation data. For the PICP values, the "low" profile exhibits lower coverage (PICP: 0.853), while the "medium" (PICP: 0.907) and "high" profile (PICP: 0.922) display higher coverage compared to the validation set. For the MPIW values, the "low" profile shows improved values (MPIW: 98.2), while the "medium" (MPIW: 151) and especially the "high" profile (MPIW: 242) exhibit larger prediction interval ranges compared to the validation set. The MRPIW for the test data follows the same trend as the validation data, showing only slight differences ($<2.5\%$). These results were deemed acceptable, permitting a more in-depth analysis of uncertainty profiles for a selected group of process activities
\begin{table}[h]
\begin{center}
\begin{minipage}{240pt}
\caption{Evaluation Metrics for the Quantile Regression Forest for Uncertainty Profiles of Test Data  (Uncertainty)}\label{Table_Model_Evaluation_per_Profile_Uncertainty_Test}
\begin{tabular*}{240pt}{@{\extracolsep{\fill}}lrrr@{\extracolsep{\fill}}}
\toprule%
Data Set            & PICP      & MPIW      & MRPIW \\
\midrule  
Test Data        &           &           &  \\
\hline
Profile: "low"        & 0.853     & 98.2      & 0.992 \\
Profile: "medium"     & 0.907     & 151.0     & 1.45 \\
Profile: "high"       & 0.922     & 242.0     & 2.27 \\
\botrule
\end{tabular*}
\end{minipage}
\end{center}
\end{table}

The examination of uncertainty profiles with regards to the process activity was focused on three activities ("Edge\_Bead", "Dishing\_Press\_1" and "Plasma\_Welding") and evaluated them using a subset of the test data related to these activities. Table \ref{Table_Model_Evaluation_per_Profile_Activity_Uncertainty} provides the PICP, MPIW, and MRPIW results for the selected activities within each profile, placing the results for point predictions in the context of model uncertainty. As observed in the general evaluation (see Table \ref{Table_Model_Evaluation_per_Profile_Uncertainty_Test}), the coverage increases from the "low" to the "high" profile. However, a significant drop in coverage is noticed for the "low" profiles of "Dishing\_Press\_1" and "Plasma\_Welding", with PICP values of 0.724 and 0.831, respectively. For the "medium" profile, the coverages are within an acceptable range, lying between 0.885 for "Edge\_Bead" and 0.905 for "Plasma\_Welding". The PICP values improve for the "high" profile, with "Dishing\_Press\_1" (0.945) and "Plasma\_Welding" (0.941) showing a noticeable increase, while "Edge\_Bead" exhibits only a slight increase (0.909). When examining the mean width of prediction intervals (MPIW), each activity displays a tendency for increasing MPIW values from the "low" to the "high" profile, but with significant differences across activities. "Edge\_Bead" shows the lowest MPIW values, ranging from 27.4 for the "medium" profile to 45.1 for the "high" profile. For "Dishing\_Press\_1", these values range from 94.0 for the "low" profile to 153.0 for the "high" profile. The highest MPIW values are found for "Plasma\_Welding", with 137.0 for the "low" profile, up to 221.0 for the "high" profile. Across activities and profiles, the MRPIW values demonstrate similar tendencies as the results of the general analysis (see Table \ref{Table_Model_Evaluation_per_Profile_Uncertainty_Test}), although slight deviations are noticeable: "Edge\_Bead" shows an increased MRPIW value for the "medium" profile, "Dishing\_Press\_1" shows an increased MRPIW value for the "low" profile, and "Plasma\_Welding" shows an increased MRPIW value for the "low" profile but a decreased MRPIW value for the "high" profile.
\begin{table}[h]
\begin{center}
\begin{minipage}{\textwidth}
\caption{Evaluation Metrics for the Quantile Regression Forest for Uncertainty Profiles and Selected Activities for Test Data (Point Prediction)}\label{Table_Model_Evaluation_per_Profile_Activity_Uncertainty}
\begin{tabular*}{\textwidth}{@{\extracolsep{\fill}}l|rrr|rrr|rrr@{\extracolsep{\fill}}}
\toprule%
& \multicolumn{3}{@{}c@{}}{PICP} & \multicolumn{3}{@{}|c|@{}}{MPIW} & \multicolumn{3}{@{}c@{}}{MRPIW} \\
\cmidrule{2-4} \cmidrule{5-7} \cmidrule{8-10}

\diagbox[outerleftsep = 0.1cm, innerwidth = 2cm]{Activity}{Profile} &  
\multicolumn{1}{c}{L\footnotemark[1]} & \multicolumn{1}{c}{M\footnotemark[2]} & \multicolumn{1}{c|}{H\footnotemark[3]} & 
\multicolumn{1}{c}{L\footnotemark[1]} & \multicolumn{1}{c}{M\footnotemark[2]} & \multicolumn{1}{c|}{H\footnotemark[3]} & 
\multicolumn{1}{c}{L\footnotemark[1]} & \multicolumn{1}{c}{M\footnotemark[2]} & \multicolumn{1}{c}{H\footnotemark[3]} \\

\midrule
"Edge\_Bead"            & NA      &  0.885      & 0.909         & NA     &  27.4     &  45.1           & NA     &  1.51      &   2.26 \\
"Dishing\_Press\_1"     & 0.724   & 0.887       & 0.945         & 94.0   & 135.0     &  153.0           & 1.09   &  1.46      &   2.24 \\
"Plasma\_Welding"       & 0.831   & 0.905       & 0.941         & 137.0   & 138.0    &  221.0          & 1.04   &  1.46      &   2.05 \\
\botrule
\end{tabular*}
\footnotetext[1]{L represents the "low" profile.}
\footnotetext[2]{M represents the "medium" profile.}
\footnotetext[3]{H represents the "high" profile.}
\end{minipage}
\end{center}
\end{table}
\subsection{SHAP Analysis for Model Uncertainty } 
This subsection investigates the SHAP values to provide insights into the contribution of each feature toward the obtained prediction intervals. The analysis is conducted both on a local level, focusing on an individual instance, and on a global level, considering the overall model behavior. First, we examine the SHAP values on the local level for the prediction interval of a specific, exemplary instance. This analysis is extended by calculating the SHAP values for the point predictions (conventional random forest), the lower boundary, and the upper boundary. Such a comparison allows us to understand the differences in feature contributions for varying levels of uncertainty. 

Next, we explore SHAP values on a global level, focusing on the overall behavior of the model. This analysis includes the examination of the SHAP feature importance and the SHAP summary plots for the prediction intervals. Following this, we compare the SHAP summary plots for each uncertainty profile ("low", "medium", and "high") to identify any differences in the contribution of features across varying levels of uncertainty. Additionally, we examine the SHAP Dependence plots for selected variables per uncertainty profile to observe any trends or patterns in feature interactions across different uncertainty profiles.

\subsubsection{Local SHAP Analysis}
\begin{figure}[h]%
\centering
\includegraphics[width = 0.95\textwidth]{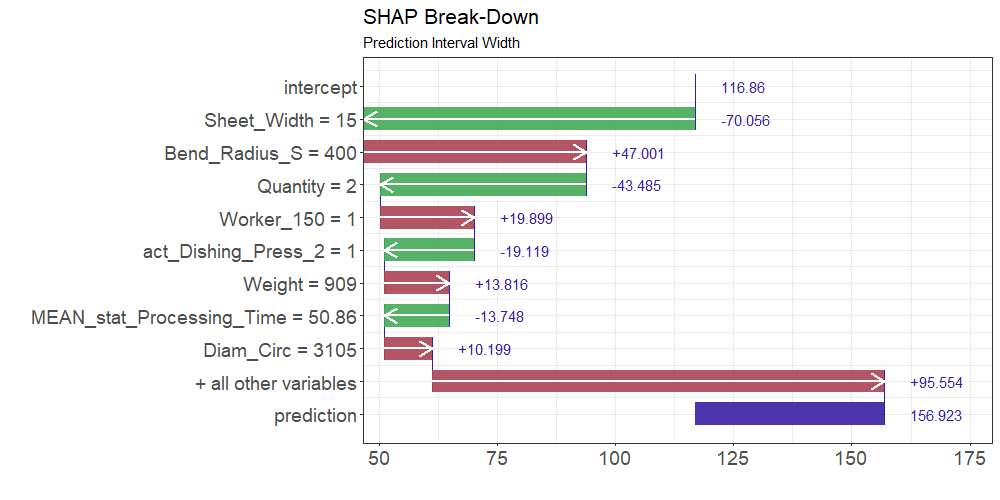}
\caption{SHAP contribution plot, depicting the impact of feature values of a specific instance to the corresponding predicted interval width. The horizontal bars depicts the relationship between variable values, as seen on the y-axis, and the corresponding impact on the final prediction, as seen on the x-axis. The y-axis starts with the intercept, followed by the eight most impactful variables based on their SHAP value for the instance, concluding with an entry representing the rest of the variables as well as the final prediction. Bars colored in red indicate an increase in interval width, while green-colored bars indicate a decrease in interval width and are supported by arrows as a visual aid. The precise SHAP-contributions, the intercept value and the final prediction are incorporated as well.}\label{SHAP_Contribution_PI_Width}
\end{figure}

Figure \ref{SHAP_Contribution_PI_Width} presents a SHAP contribution plot for the predicted interval of a selected instance from the test dataset. This specific instance belongs to the "low" profile and is associated with the activity "Dishing\_Press\_2". The plot demonstrates the impact of various variables on the prediction interval width, with variables having the highest absolute impact placed at the top of the y-axis. For the selected instance, the sheet width ("Sheet\_Width = 15") has the most significant impact on narrowing the prediction interval, reducing its width by 70.056 minutes. On the other hand, a specific bend radius ("Bend\_Radius\_S = 400") has the most substantial impact on expanding the prediction interval, increasing its width by 47.001 minutes. Due to a large number of variables, the plot only displays the top eight most impactful variables. The SHAP values of the remaining variables are summed up and incorporated as "all other variables", which contribute to widening the interval width by another 95.554 minutes.

\begin{figure}[h]%
\centering
\includegraphics[width =0.95\textwidth]{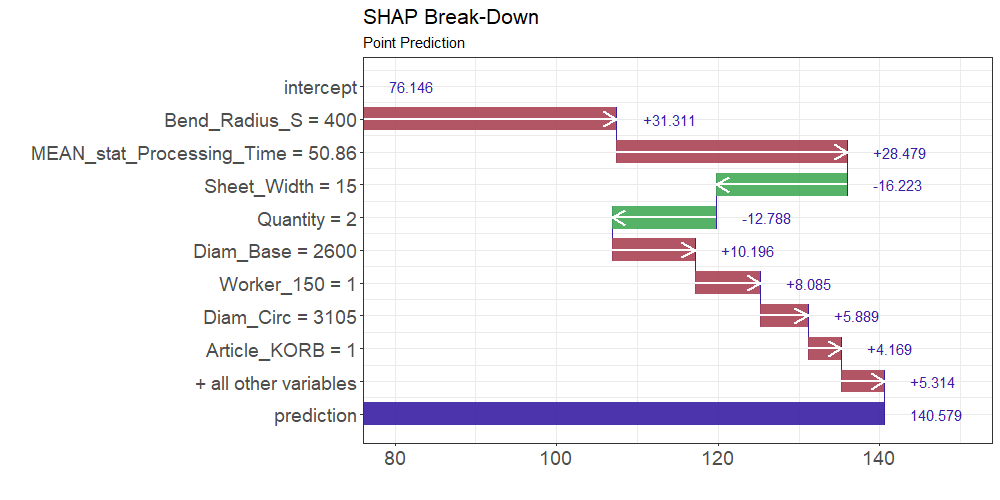}
\caption{SHAP contribution plot, in the same fashion as Figure \ref{SHAP_Contribution_PI_Width}, depicting the impact of feature values of a specific instance on the point prediction.}\label{SHAP_Contribution_Point_Prediction}
\end{figure}

\begin{figure}[h]%
\centering
\includegraphics[width = 0.95\textwidth]{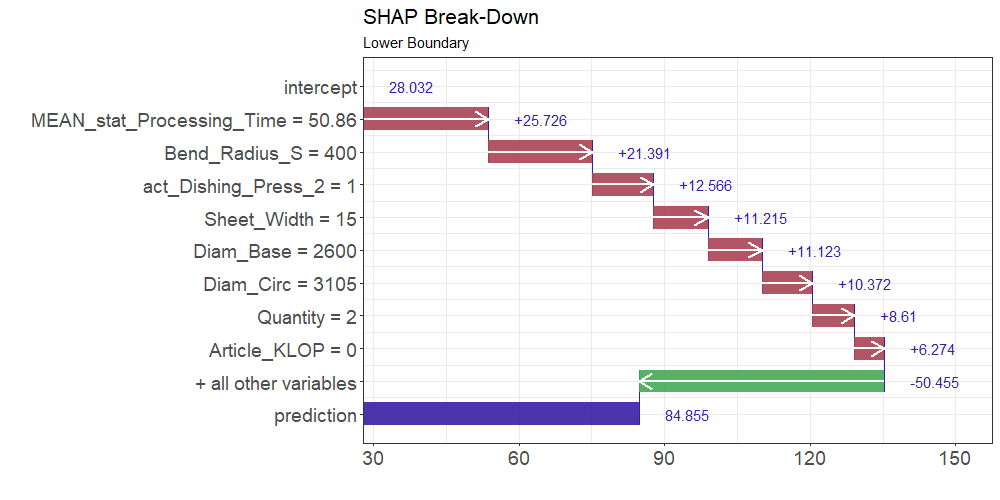}
\caption{SHAP contribution plot, in the same fashion as Figure \ref{SHAP_Contribution_PI_Width}, depicting the impact of feature values of a specific instance on the lower boundary of the prediction interval}\label{SHAP_Contribution_Lower_Boundary}
\end{figure}

\begin{figure}[h]%
\centering
\includegraphics[width = .95\textwidth]{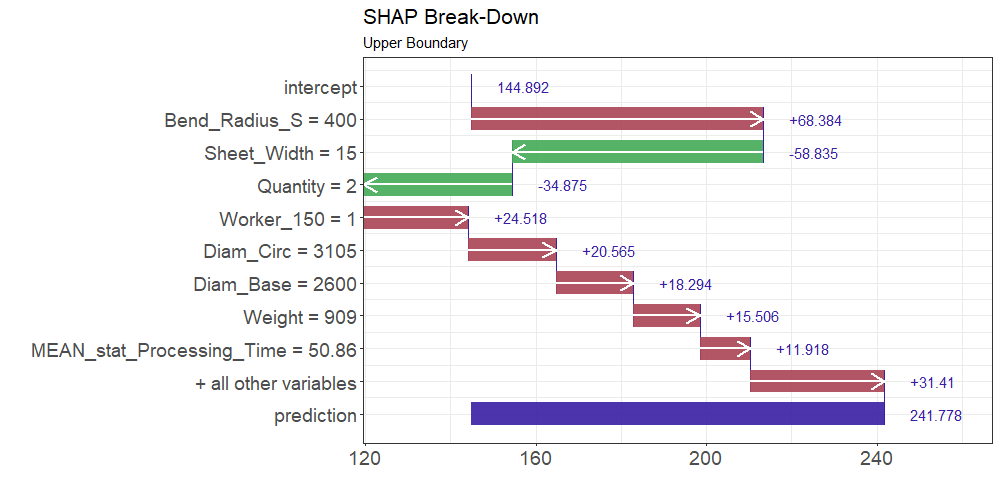}
\caption{SHAP contribution plot, in the same fashion as Figure \ref{SHAP_Contribution_PI_Width}, depicting the impact of feature values of a specific instance on the upper boundary of the prediction interval.}\label{SHAP_Contribution_Upper_Boundary}
\end{figure}

Figures \ref{SHAP_Contribution_Point_Prediction}, \ref{SHAP_Contribution_Lower_Boundary}, and \ref{SHAP_Contribution_Upper_Boundary} display the SHAP contribution plots for the point prediction, lower boundary, and upper boundary of the prediction interval, respectively, for the same selected instance as examined in Figure \ref{SHAP_Contribution_PI_Width}. Due to different intercept values, there are significant differences in the magnitude of SHAP values across the three plots, especially between the lower and upper boundaries. For the conventional point prediction (see Figure \ref{SHAP_Contribution_Point_Prediction}), the "Bend\_Radius\_S" value exhibits the highest SHAP contribution, increasing the predicted duration of the processing time by 31.311 minutes, while "Sheet\_Width" value decreases the prediction by 16.223 minutes. Although a similar trend to the SHAP values for the prediction interval can be observed (see Figure \ref{SHAP_Contribution_PI_Width}), the "Bend\_Radius\_S" variable has a stronger influence on the point prediction than on the prediction interval in, while the impact of the "Sheet\_Width" variable is comparatively smaller on the point prediction.

For the lower boundary (see Figure \ref{SHAP_Contribution_Lower_Boundary}), an intercept of 28.032 is documented, with the "MEAN\_stat\_Processing\_Time" value having the most significant impact on the prediction, increasing the predicted value by 25.726. In contrast, for the upper boundary (see Figure \ref{SHAP_Contribution_Upper_Boundary}), the intercept of 144.892 is impacted most by the "Bend\_Radius\_S" variable, increasing the predicted value by 68.384 – more than twice as much as the most important variable of the lower boundary. Furthermore, discrepancies in the ranking of variables indicate differences in the influence of variable values on the predicted subject. For instance, the "Sheet\_Width" and "Quantity" values increase the final result of the lower boundary while decreasing the result of the upper boundary. This observation suggests a narrowing of the prediction interval, leading to a decrease in model uncertainty, which is also confirmed by the SHAP contribution plot for the PI (see Figure \ref{SHAP_Contribution_PI_Width}).

Based on our preliminary assessment conducted in collaboration with process experts, the SHAP analysis significantly enhances the comprehensive understanding of a ML's decision-making process for individual instances. This enables experts to identify and validate crucial features that impact specific predictions. Gaining such insights not only aids in identifying potential biases or unforeseen feature contributions, which may require model adjustments or data preprocessing, but also promotes better communication and trust among stakeholders with diverse technical backgrounds. Moreover, the interpretability provided by the SHAP analysis empowers domain experts to more effectively collaborate with data scientists in refining the model, ensuring that it aligns with domain knowledge and real-world expectations, ultimately leading to more accurate and reliable predictions.  

\subsubsection{Global SHAP Analysis}
\begin{figure}[h]%
\centering
\includegraphics[width = 0.9\textwidth]{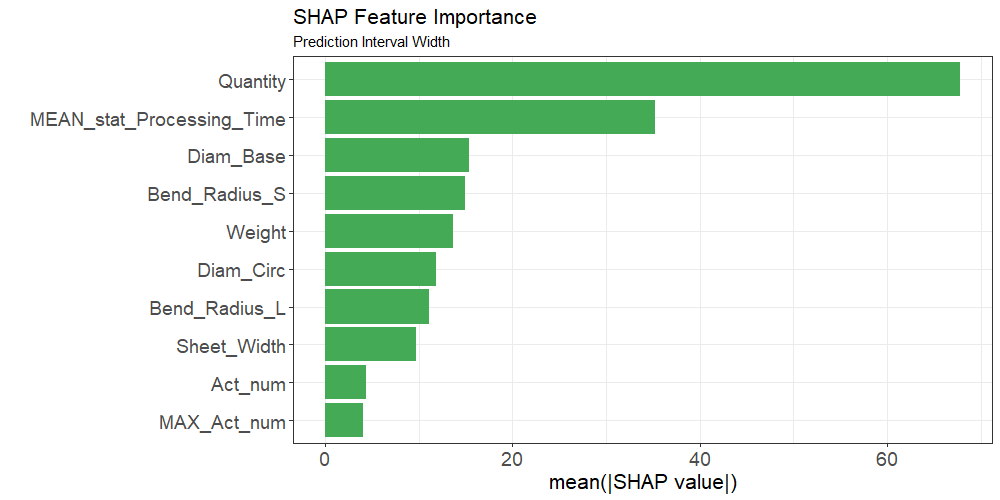}
\caption{SHAP feature importance plot, depicting the impact of feature values on the final model prediction on a global level. This plot visualizes the ten most important variables in descending order, as seen on the y-axis, and their corresponding impact via the length of horizontal bars, as seen on the x-axis. The importance of a variable is calculated by averaging the absolute SHAP values documented for the corresponding variable for the test data set.}\label{SHAP_FI}
\end{figure}

Figure \ref{SHAP_FI} presents the SHAP feature importance, generated for the test dataset, and illustrates the model behavior on a global level. Specifically, the top ten variables with the highest absolute impact on the prediction of interval widths are shown, with the production quantity ("Quantity") and the average statistical duration ("MEAN\_stat\_Processing\_Time") being the most impactful variables.

\begin{figure}[h]%
\centering
\includegraphics[width = 0.9\textwidth]{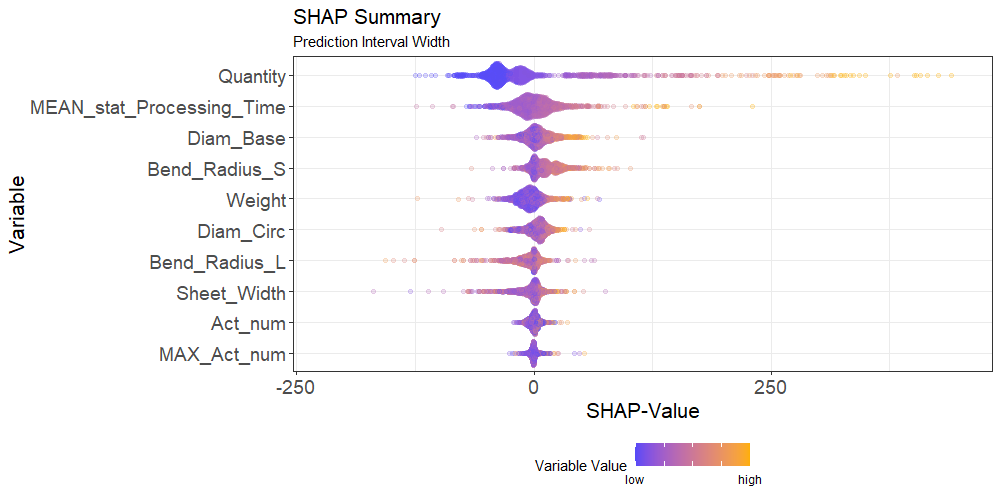}
\caption{SHAP summary plot for the ten most impactful variables, depicting the relationship between variable values and corresponding SHAP values. The variables are represented on the y-axis, SHAP values on the x-axis. For the visualization of the distribution of SHAP values, a mixed approach of beeswarm and violin plot was chosen: Each point represents an instance, with the color of the point depicting the relative variable value, its position on the x-axis representing the SHAP value and its position on the y-axis within the bounds of the variable complying with the density of the area.}\label{SHAP_Summary_PI_width}
\end{figure}

For a more detailed analysis, Figure \ref{SHAP_Summary_PI_width} is a SHAP summary plot that enables the examination of the relationship between feature values and their corresponding impact on the predicted interval width. For each variable depicted on the y-axis, each point represents its feature value through its color and the corresponding SHAP value via its position on the x-axis. For instance, low values of the "Quantity" variable are associated mainly with negative SHAP values, thus reducing the prediction interval, while high values of the "Quantity" variable are associated with positive SHAP values, leading to the expansion of prediction intervals. Due to the number of variables in the dataset, only the top 10 most important variables are displayed in descending order, with the "Quantity" variable being the most impactful feature.

The utilization of global SHAP explanations, encompassing SHAP summary plots, SHAP feature importance, and SHAP Dependence plots, confers several significant advantages to process experts in understanding complex ML models. By elucidating the intricate relationships between input features and model predictions, these explanations furnish practitioners with a holistic understanding of the model's behavior, which subsequently facilitates model optimization. Furthermore, SHAP summary plots provide an aggregated view of feature importance across all instances, enabling the identification of overarching trends and the relative significance of each feature in the model. In addition, SHAP feature importance rankings offer a granular, instance-level perspective of feature contributions, thus empowering process experts to discern the salient features driving specific predictions. Lastly, SHAP Dependence plots explicate the interaction effects between features, thereby unveiling potential synergies or redundancies among them. Collectively, these global SHAP explanations not only enhance process experts' capacity to interpret and validate ML models but also foster transparency, trust, and comprehensibility in the context of data-driven decision-making.

\subsubsection{SHAP Analysis of Uncertainty Profiles}
\begin{figure}[h]%
\centering
\includegraphics[width = 0.9\textwidth]{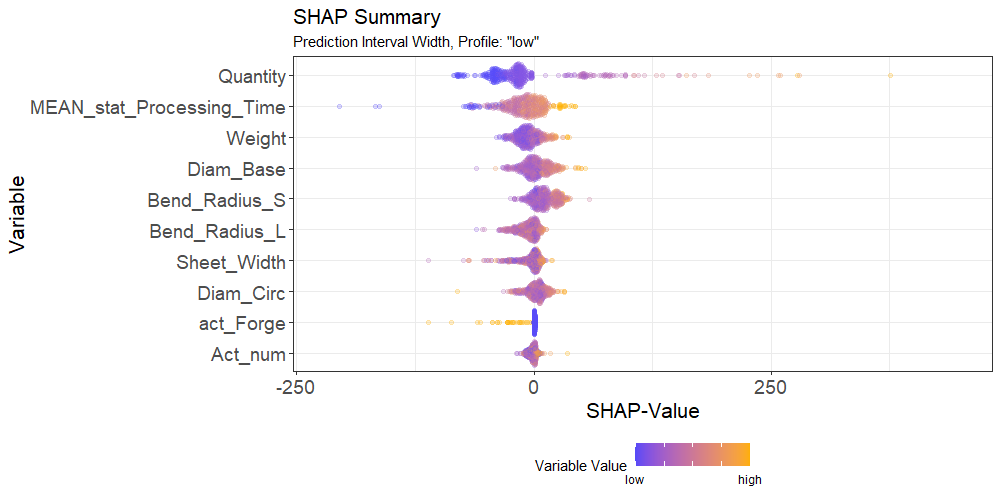}
\caption{SHAP summary plot for the profile "low". The same approach as in Figure \ref{SHAP_Summary_PI_width} was used, but the data was restricted to instances pertaining to the "low" uncertainty group.}\label{SHAP_Summary_low}
\end{figure}

\begin{figure}[h]%
\centering
\includegraphics[width = 0.9\textwidth]{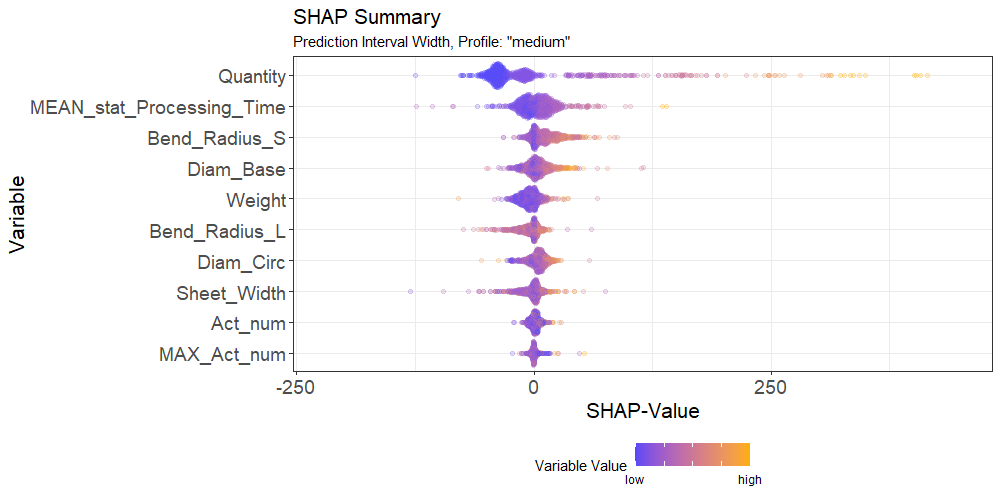}
\caption{SHAP summary plot for the profile "medium". The same approach as in Figure \ref{SHAP_Summary_PI_width} was used, but the data was restricted to instances pertaining to the "medium" uncertainty group.}\label{SHAP_Summary_medium}
\end{figure}

\begin{figure}[h]%
\centering
\includegraphics[width = 0.9\textwidth]{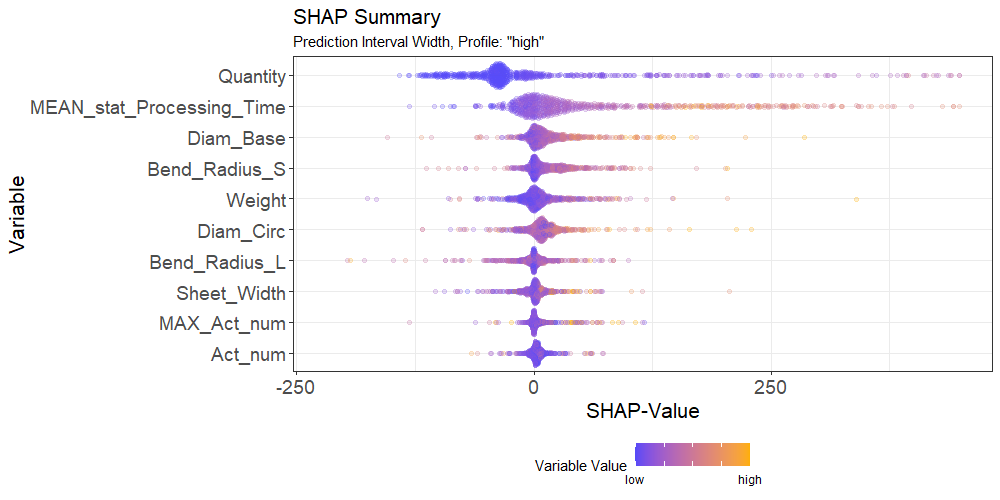}
\caption{SHAP summary plot for the profile "high". The same approach as in Figure \ref{SHAP_Summary_PI_width} was used, but the data was restricted to instances pertaining to the "high" uncertainty group.}\label{SHAP_Summary_high}
\end{figure}

In the realm of uncertainty quantification (UQ), it has been noted in Section \ref{Uncertainty_Profile_Construction_and_Evaluation} that there exist discrepancies in model behavior among distinct uncertainty profiles. To examine these variations, Figures \ref{SHAP_Summary_low}, \ref{SHAP_Summary_medium}, and \ref{SHAP_Summary_high} illustrate the SHAP summary plots corresponding to the "low," "medium," and "high" profiles, respectively. The subsequent observations delineate salient disparities in model behavior for each individual profile. Although the "Quantity" and "MEAN\_stat\_Processing\_Time" variables emerge as the most influential features, a divergent ranking of the most consequential variables is discernible across profiles. For instance, the significance of the processed material's weight ("Weight") for predicting interval widths is more pronounced in the "low" profile, where it ranks as the third most crucial variable, compared to the "medium" or "high" profiles, in which it occupies the fifth position.

Additionally, disparities in the shape and dispersion of the SHAP value distributions for various variables are observable. In the "low" profile, the distribution for the "MEAN\_stat\_Processing\_Time" variable exhibits left skewness, while the "Diam\_Base" variable (representing the product's base diameter) demonstrates right skewness. Comparing the "MEAN\_stat\_Processing\_Time" variable distribution across uncertainty profiles reveals minimal skewness in the "medium" profile and right skewness in the "high" profile. Moreover, it is evident that the "high" profile's distributions exhibit a substantial spread, followed by the "medium" profile, whereas the "low" profile manifests the most minimal dispersion in SHAP values.

\begin{figure}[h]%
\centering
\includegraphics[width = 0.9\textwidth]{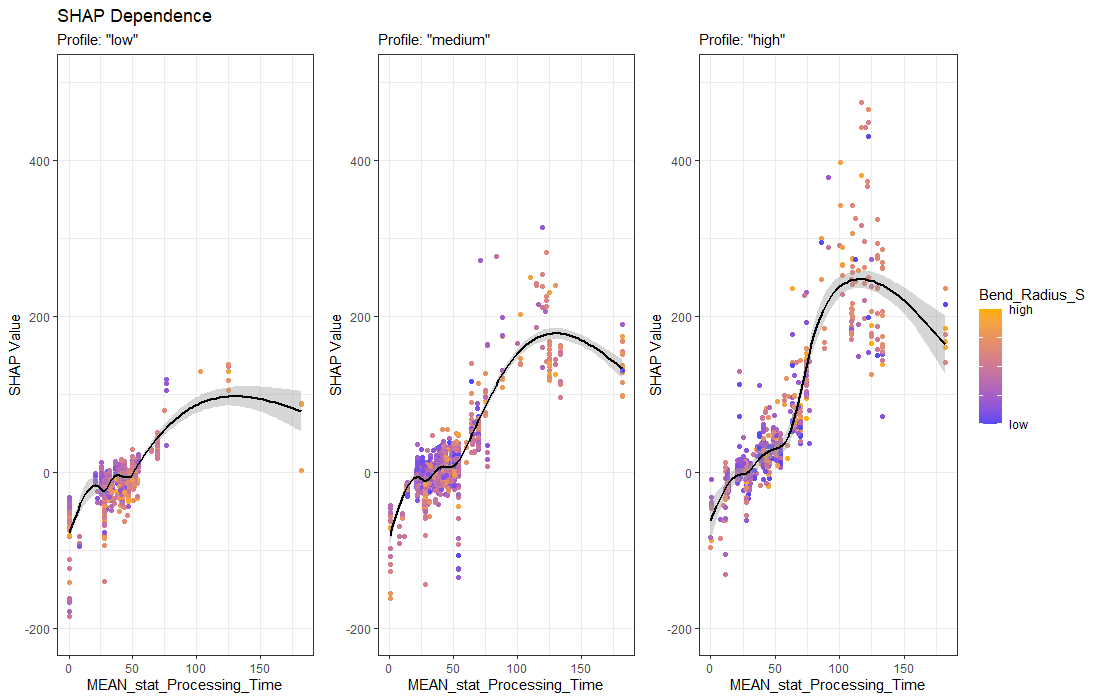}
\caption{SHAP dependence plot for the variable "MEAN\_stat\_Processing\_Time", with the secondary variable "Bend\_Radius\_S" for the "low", "medium" and "high" uncertainty profile. Each point represents the relationship between the "MEAN\_stat\_Processing\_Time" value of an instance, as seen on the x-axis, the corresponding SHAP value, as seen on the y-axis, and the corresponding relative "Bend\_Radius\_S" value, provided via color coding. A black smoothing curve, calculated via a general additive model, provide a visual aid for each plot. }\label{SHAP_Dependence}
\end{figure}

The examination of specific variables and their impact on the predicted interval can be further explored using SHAP dependence plots. This visualization supports the analysis of model behavior by depicting the relationship between values of selected variables and corresponding SHAP values. Figure \ref{SHAP_Dependence} provides three SHAP dependence plots for the "MEAN\_stat\_Processing\_Time" and "Bend\_Radius\_S" variables, one for each uncertainty profile. These plots provide the values of the "MEAN\_stat\_Processing\_Time" variable on the x-axis and the values of the "Bend\_Radius\_S" variable via color-coding while the corresponding SHAP values are provided on the y-axis. Each plot integrates a smoothing curve, calculated via generalized additive models, in order to mitigate the observed overplotting and identify trends of model behavior. All of the plots indicate a tendency to increase the SHAP value with higher "MEAN\_stat\_Processing\_Time" values, although some differences in the distribution of "MEAN\_stat\_Processing\_Time" values can be observed between profiles. The "low" profile documents only few "MEAN\_stat\_Processing\_Time" values above 100, followed by the "medium" profile and lastly the "high" profile, showing the highest amount of "MEAN\_stat\_Processing\_Time" values. The magnitude of SHAP values increases for "MEAN\_stat\_Processing\_Time" values within said interval from the "low" to the "high" profile as well, as indicated by the smoothing curves.  

The investigation of specific features and their influence on the predicted interval width can be further delved into through the employment of SHAP dependence plots. These visualizations facilitate the analysis of model behavior by delineating the interactions between the values of chosen features and their corresponding SHAP values. Figure \ref{SHAP_Dependence} displays three SHAP Dependence plots for the "MEAN\_stat\_Processing\_Time" and "Bend\_Radius\_S" variables, one corresponding to each uncertainty profile. These plots feature the "MEAN\_stat\_Processing\_Time" variable values on the x-axis, the "Bend\_Radius\_S" variable values represented via color-coding, and the associated SHAP values on the y-axis. To alleviate the observed overplotting and discern trends in model behavior, each plot incorporates a smoothing curve calculated using generalized additive models. As accentuated by the smoothing curves, all plots exhibit an inclination for the SHAP value to increase with higher "MEAN\_stat\_Processing\_Time" values, although certain disparities in the distribution of "MEAN\_stat\_Processing\_Time" values are discernible among profiles. The "low" profile records only a few "MEAN\_stat\_Processing\_Time" values exceeding 100, followed by the "medium" profile, and finally, the "high" profile, which displays the greatest number of "MEAN\_stat\_Processing\_Time" values. The magnitude of SHAP values also escalates for "MEAN\_stat\_Processing\_Time" values within the aforementioned interval, progressing from the "low" to the "high" profile, as indicated by the smoothing curves.

Incorporating global SHAP analysis for subsets of data with distinct uncertainty profiles offers valuable insights into the model's behavior under varying levels of uncertainty, which can be instrumental in enhancing its robustness and generalizability. Analyzing these subsets allows process experts to identify features that contribute significantly to predictions in specific uncertainty contexts, thereby shedding light on potential areas for model refinement and improvement. Moreover, this targeted examination of data subsets may reveal disparities in feature importance and interactions across different uncertainty profiles, elucidating the model's sensitivity to diverse input conditions. Consequently, such an approach bolsters the understanding of the model's limitations and strengths and fosters informed decision-making for various uncertainty scenarios, ultimately contributing to a more reliable and trustworthy machine learning model. By dissecting the model's performance across distinct uncertainty profiles, process experts can also establish tailored mitigation strategies for each scenario, enhancing the model's efficacy in real-world applications.

\section{Discussion} \label{Discussion}
\subsection{Implications for Domain Experts}
The utilization of a multi-stage machine learning approach that integrates uncertainty awareness and explainability holds significant implications for decision-makers across diverse business operations. Through the use of machine learning models that account for uncertainty, experts in a given field can gain a deeper understanding of potential outcomes and their associated variability. This enhanced comprehension can facilitate more efficent decision-making and ultimately lead to improved risk management. The heightened awareness of uncertainty leads to improved operational processes in organizations, promoting a culture of decision-making based on data, which ultimately results in increased efficiency and effectiveness. The integration of ML explanation components in the decision-making process offers the fundamental benefit of being able to discern the fundamental factors that contribute to predictive uncertainty. This factor provides decision-makers with the ability to concentrate their endeavors on mitigating the pertinent origins of unpredictability, thereby enhancing the resilience and trustworthiness of the decision-making mechanism. 

Moreover, the knowledge acquired from our uncertainty-aware explainable approach can be utilized to enhance resource allocation, facilitating organizations to give priority to resources in domains with significant volatility and alleviate related risks. This solution can also provide benefits for domain experts in the areas of strategic planning and organizational adaptability. By comprehending the magnitude of uncertainties, individuals can formulate more reliable and adaptable strategic plans that correspond with the objectives of the organization and ensure sustained prosperity. In addition, the capacity to measure and clarify uncertainties provides professionals with the necessary assets to adapt to evolving circumstances and address possible interruptions, augmenting the competitiveness of the enterprise in a dynamic commercial setting. 

\subsection{Theoretical/Scientific Implications}
The utilization of our proposed approach that integrates uncertainty awareness and explainability has noteworthy theoretical and scientific implications for the domains of prescriptive analytics, operations research, and artificial intelligence. This study makes a contribution to the advancement of scientific knowledge by addressing gaps in the existing literature. Specifically, it emphasizes the significance of integrating technical production parameters, producing machine learning outputs that account for uncertainties, and elucidating the origins of such uncertainties. The incorporation of these components within the decision-making framework has the potential to enhance the efficacy of the model, produce more resilient optimization results, and foster a deeper comprehension of intricacies inherent in practical scenarios.

From a methodological perspective, the proposed approach expands the use of machine learning methods, such as QRF and SHAP, to generate prediction intervals and attribute uncertainty to particular input features. The progress made in this field not only facilitates a more thorough comprehension of the inherent uncertainties in problems related to operations research but also lays the groundwork for the creation of novel methodologies and techniques that further augment the amalgamation of uncertainty and explication in models used for optimization. Consequently, forthcoming studies may utilize these methodological advancements to develop novel approaches that tackle a diverse range of intricate commercial challenges.

Finally, by illustrating its applicability to real production planning scenarios, the proposed method makes a significant addition to the scientific community. This use case serves as a proof-of-concept, demonstrating how the multi-stage machine learning strategy is effective at managing uncertainty and delivering useful insights. The successful application of the suggested strategy in a practical setting may inspire additional investigation and study in related areas, fostering interdisciplinary cooperation and encouraging the creation of new theories, methodologies, and applications that advance scientific understanding generally.

\subsection{Threats to Validity}
Although the proposed methodology exhibits encouraging outcomes in the domain of predictive process monitoring, it is imperative to recognize plausible threats to the study's validity. The identification of these potential threats facilitates a more comprehensive comprehension of the limitations inherent in the study, thereby promoting the pursuit of additional research endeavors aimed at mitigating them.

The validity of the findings can be significantly influenced by the quality and representativeness of the data utilized in this study. The case study's findings might not be generalizable if the data used don't accurately reflect the real-world scenario or if they have biases, inconsistencies, or errors. Moreover, it is imperative to have an adequately large sample size to mitigate the impact of random variations or anomalies on the results. The validity of the study may be impacted by the assumptions made during the development of the QRF and Kernel SHAP models. It is important to note that the assumptions regarding the underlying distribution of the data and the interactions between variables may not be applicable in all scenarios. Thus, the efficacy of the suggested methodology may exhibit variability contingent upon these aforementioned factors. The validity of the study may be compromised if the suggested methodology lacks robustness towards variations in the input data or exhibits excessive sensitivity to particular parameter configurations. Conducting a sensitivity analysis is of utmost importance in assessing the resilience of the model to variations in input data and hyperparameters. It would enable to ascertain whether the performance of the approach remains consistent across various conditions or if it is vulnerable to slight disturbances.

Although SHAP assists in offering understanding regarding the origins of uncertainty, the capacity for interpretation and clarification of these explanations may remain restricted. The comprehensibility of the factors that contribute to uncertainty may be impeded by complicated interrelationships among variables or the existence of data with a high number of dimensions, which can pose challenges for experts in the field. Subsequent investigations ought to prioritize the development of explanations that are more practical and comprehensible, thereby facilitating effective communication with relevant parties. Through the mitigation of these potential sources of error, forthcoming investigations can enhance the suggested methodology and augment the overall comprehension of UQ and XAI within predictive process monitoring.

\section{Related Work} \label{Background_and_Related_Work}
The overarching objective of OR is to enhance methodologies for making judicious and efficacious managerial decisions; thus, it is paramount that information systems and decision support tools are intricately and coherently integrated \cite{simon1997future}. Due to the relative scarcity of data and limitations in computing power, operations management research has predominantly relied on models based on microeconomic theory, game theory, optimization, and stochastic models to generate strategic insights into how firms should operate \cite{mivsic2020data}. The amelioration of these limitations, in conjunction with significant algorithmic advancements, has established AI as a crucial enabler within the context of OR.  AI has the capability to narrow down the range of possible decisions, which can facilitate the use of data-driven analytics in optimizing operations. For instance, in the context of scheduling, AI fosters heightened awareness and comprehension of the underlying processes, ultimately enabling more streamlined and efficient operations \cite{isaksson2018impact}.

The interplay between these two domains is bidirectional, as machine learning models used in AI are typically trained by solving optimization problems, which are integral to an operations researcher’s toolbox, and in turn, AI techniques are employed in OR to predict key parameters and develop heuristics for solving complex optimization problems \cite{bennett2006interplay}. In the realm of conditional-stochastic optimization, Bertsimas and Kallus explored the adaptation of predictive analytics methods to estimate conditionally expected costs for any input, addressing the challenge of minimizing uncertain costs given incomplete information \cite{bertsimas2020predictive}. Their work demonstrates the potential of leveraging predictive analytics techniques to tackle more intricate optimization problems, paving the way for novel applications in prescriptive analytics. Bengio et al (2021) investigate the synergistic potential of machine learning and combinatorial optimization, advocating for a methodology that treats optimization problems as data points to identify relevant problem distributions and enhance decision-making beyond traditionally handcrafted heuristics \cite{bengio2021machine}.

\begin{figure}[h]%
\centering
\includegraphics[width = 0.9\textwidth]{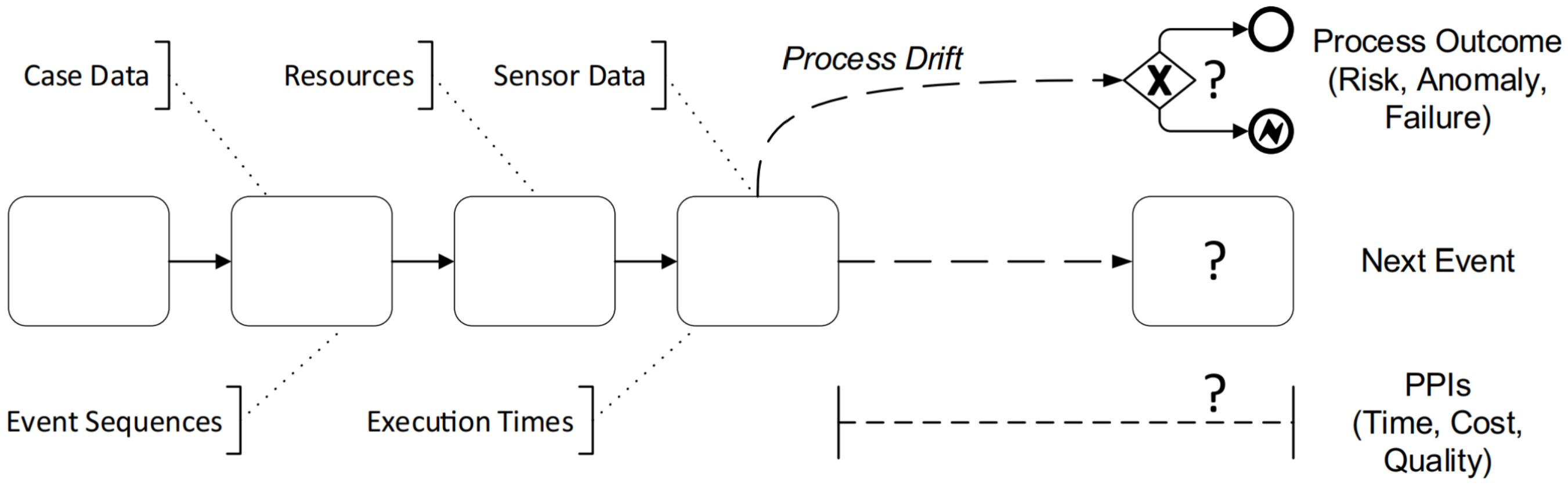}
\caption{Overview of predictive process analytics \cite{rehse2019towards}}\label{Figure_Overview_Predictive_Process_Analytics}
\end{figure}

In this study, we focus on a specific ml problem, namely predictive process monitoring, which is a technique within the broader field of process mining that includes process discovery, conformance checking, and process enhancement \cite{van2012_process_manifesto}. Predictive process monitoring leverages historical execution data to provide users with predictions about a target of interest for a given process execution \cite{maggi2014_predictive_process_monitoring}. Process mining encompasses a set of techniques aimed at extracting valuable insights from data generated by process-aware information systems during process execution. It serves as an intermediary between process science (including operations research) and data science (encompassing fields such as predictive and prescriptive analytics), offering methods for data-driven process analysis \cite{van2022process}. As illustrated in Figure \ref{Figure_Overview_Predictive_Process_Analytics} and presented in \cite{rehse2019towards}, there are three central prediction tasks based on the target of interest and its characteristics: process outcome prediction \cite{teinemaa2019_outcome_prediction}, next event prediction \cite{tax2017_next_event, evermann2017_next_event}, and remaining time prediction \cite{verenich2019_remaining_time, teinemaa2018_time_prediction}.

Numerous review articles have been published on the subject of predictive process monitoring. For example, \cite{di2018_which_suits_best} classified 51 process prediction methods based on their prediction targets using a value-driven framework. These methods exhibited different prediction architectures and were categorized into various categories, including categorical outcome, costs, inter-case metrics, risk, sequence of values, and time. \cite{teinemaa2019_outcome_prediction} conducted a systematic review and proposed a taxonomy for outcome-oriented predictive process monitoring. The authors identified and compared 14 relevant papers based on several criteria, including classification algorithm, filtering, prefix extraction, sequence encoding, and trace bucketing. Additionally, an experimental evaluation capturing the impact of different qualitative criteria was conducted using the authors' own implementation. \cite{verenich2019_remaining_time} conducted a survey on methods for predicting remaining time in business processes, examining and comparing 25 relevant papers published between 2008 and 2017 based on criteria such as application domain, input data, prediction algorithm, and process awareness. A quantitative comparison was performed via a benchmark of 16 remaining time prediction methods on various publicly available datasets.

In practical applications, black-box ML algorithms are increasingly efficient in achieving a level of accuracy for predictive process monitoring unattainable by conventional, inherently interpretable ML approaches \cite{neu2022systematic}. However, these opaque methods often lack explanations for their working mechanisms, forcing users to rely on transparent, albeit often suboptimal, models \cite{arrieta2020explainable}. Providing explanations is an effective strategy for fostering the acceptance of predictions offered by intelligent systems \cite{mehdiyev2021explainable}. As a result, explainable artificial intelligence (XAI) has emerged as a promising research domain focused on facilitating collaboration between AI-based systems and human users by enhancing the transparency of the underlying opaque algorithms \cite{guidotti2018survey}. Numerous scholarly investigations have conducted comprehensive systematic reviews of existing literature, delving deeply into the multifaceted dimensions of explanation techniques within the context of artificial intelligence and machine learning \cite{emmert2020explainable}. 

These studies have explored the dichotomy between local and global methods, considering their distinct approaches and implications \cite{adadi2018peeking}. Moreover, the relationship between explanatory techniques and their corresponding models has been thoroughly investigated, differentiating between model-specific and model-agnostic approaches \cite{angelov2021explainable}. A crucial aspect of these studies involves examining various stakeholders in both the design and application stages of explanation techniques, acknowledging their diverse perspectives and requirements \cite{arrieta2020explainable}. Concurrently, researchers have strived to discern the overarching objectives of these explanatory mechanisms, uncovering the underlying motivations and intentions that drive their development and implementation \cite{mehdiyev2021explainable}. Furthermore, the interdisciplinary nature of these inquiries has facilitated the integration of cognitive and social science perspectives, fostering a more profound understanding of the human factors influencing the comprehension and interpretation of explanations offered by machine learning models \cite{miller2019explanation}. Contextual factors have also been taken into account, emphasizing the significance of situational elements in determining the relevance and efficacy of explanatory techniques. Finally, the evaluation mechanisms employed to assess the effectiveness and utility of explanation methods have been rigorously analyzed, establishing essential criteria and benchmarks that facilitate a stringent and objective assessment of their performance \cite{vilone2021notions, van2021evaluating}. Several studies have implemented XAI approaches for predictive process monitoring problems, primarily focusing on post-hoc explanation techniques \cite{harl2020explainable,stevens2022quantifying,velmurugan2021evaluating,mehdiyev2020local,mehdiyev2020prescriptive}.

UQ constitutes another emerging domain in ML research, emphasizing the estimation and effective communication of uncertainty inherent in predictions generated by ML models. UQ can be perceived as a supplementary form of transparency, which enhances the understandability of resolutions for decision-making tasks, an aspect potentially not addressed sufficiently by other means \cite{bhatt2021uncertainty}. When appropriately calibrated and effectively communicated, uncertainty augments stakeholders' ability to determine when to rely on model predictions, thereby improving the efficiency of decision automation or support systems. By systematically quantifying and incorporating uncertainty into the analysis, UQ approaches facilitate more robust and reliable optimization strategies and decision-making frameworks, thus promoting well-informed decision-making under ambiguous conditions \cite{ghanem2017handbook}. As a result, the application of UQ methods has the potential to significantly improve the effectiveness and resilience of solutions across various disciplines, including engineering, finance, and environmental management \cite{smith2013uncertainty}. Recently, a few approaches have been proposed for addressing uncertainty in predictive process monitoring \cite{weytjens2022learning, shoush2022intervene}.

The bidirectional integration of UQ and XAI, concentrating on elucidating the sources of uncertainties and examining the uncertainties inherent in explanations, has largely been overlooked in academic research. It is only recently that a limited number of studies have begun to focus on integrating approaches from these two research domains \cite{slack2021reliable, antoran2020getting, moosbauer2021explaining}. This article is among these pioneering efforts, aiming to contribute to this burgeoning field by making model uncertainties comprehensible to domain experts in the context of a predictive process monitoring problem. Furthermore, to the best of our knowledge, our study represents the first of its kind to integrate UQ and XAI in the context of predictive process monitoring problems.

\section{Conclusion} \label{Conclusion}
This study has introduced a comprehensive, multi-stage machine learning approach that adeptly integrates information systems and artificial intelligence techniques within an operations research framework to address common limitations in extant solutions. These limitations encompass the lack of data-driven estimation for essential production parameters, the sole generation of point forecasts without considering model uncertainty, and the absence of explanations regarding the origins of such uncertainty. By employing QRF for generating interval predictions and both local and global variants of SHAP for predictive process monitoring, the research aims to enrich the understanding of the domain experts. This underscores the significance of incorporating technical production parameters, producing machine learning outputs that accommodate uncertainties, and elucidating the sources of these uncertainties. The practical applicability of the proposed method was substantiated through a real-world production planning case study, illustrating its efficacy in handling uncertainty and delivering valuable insights. The successful application of the multi-stage machine learning strategy in a pragmatic context highlights its potential for improving decision-making processes and promotes further interdisciplinary collaboration, as well as the development of novel theories, methodologies, and applications in related fields.

\section*{Compliance with Ethical Standards} \label{Compliance with Ethical Standards}

\bmhead{Funding} This research was funded in part by the German Federal Ministry of Education and Research under grant number 01IS21006B (project ExPro).

\bmhead{Conflict of interest} The authors declare that they have no conflict of interest.

\bmhead{Ethical approval} This article does not contain any studies with human participants or animals performed by any of the authors.

\bibliography{sn-bibliography}

\end{document}